\documentclass{article}

\usepackage[final]{corl_2026} 

\usepackage{amsmath}
\usepackage{amssymb}
\usepackage{mathtools}
\usepackage{amsthm}
\usepackage{booktabs}

\usepackage{graphicx}
\usepackage{caption}
\usepackage{bm}
\newcommand{\cmark}{\ding{51}} 
\newcommand{\xmark}{\ding{55}} 
\usepackage[table]{xcolor}
\usepackage{pifont}
\usepackage{multirow}
\usepackage{wrapfig} 

\usepackage{makecell}
\usepackage{tabularx}
\usepackage{float}
\usepackage{array}
\usepackage{enumitem}
\usepackage{placeins}

\title{Schrödinger's Navigator: Imagining an Ensemble of Futures for Zero-Shot Object Navigation}

\author{%
\begin{tabular}{c}
\textbf{Yu He$^{1,4,6}$, Da Huang$^{2,4}$, Zhenyang Liu$^{1,4,6}$, Zixiao Gu$^{2,4,6}$, Qiang Sun$^{3}$}\\
\textbf{Guangnan Ye$^{1,4,\dagger}$, Yanwei Fu$^{1,4,7,\dagger,\ddagger}$, Yu-Gang Jiang$^{1}$}\\[8pt]
{\normalfont $^{1}$Fudan University \quad
$^{2}$Shanghai Jiao Tong University}\\
{\normalfont $^{3}$Shanghai University of International Business and Economics}\\
{\normalfont $^{4}$Shanghai Innovation Institute \quad
$^{6}$ActiMind}\\
{\normalfont $^{7}$Shanghai Key Laboratory of Multimodal Embodied AI}\\[8pt]
{\normalfont $^{\dagger}$Corresponding authors \quad $^{\ddagger}$Project leader}\\[8pt]
{\normalfont Project Page: \protect\url{https://heyu322.github.io/Schrodinger-Navigator.github.io/}}
\end{tabular}%
}

\begin{document}
\maketitle

\begin{figure}[!ht]

  \centering

  \includegraphics[width=\textwidth]{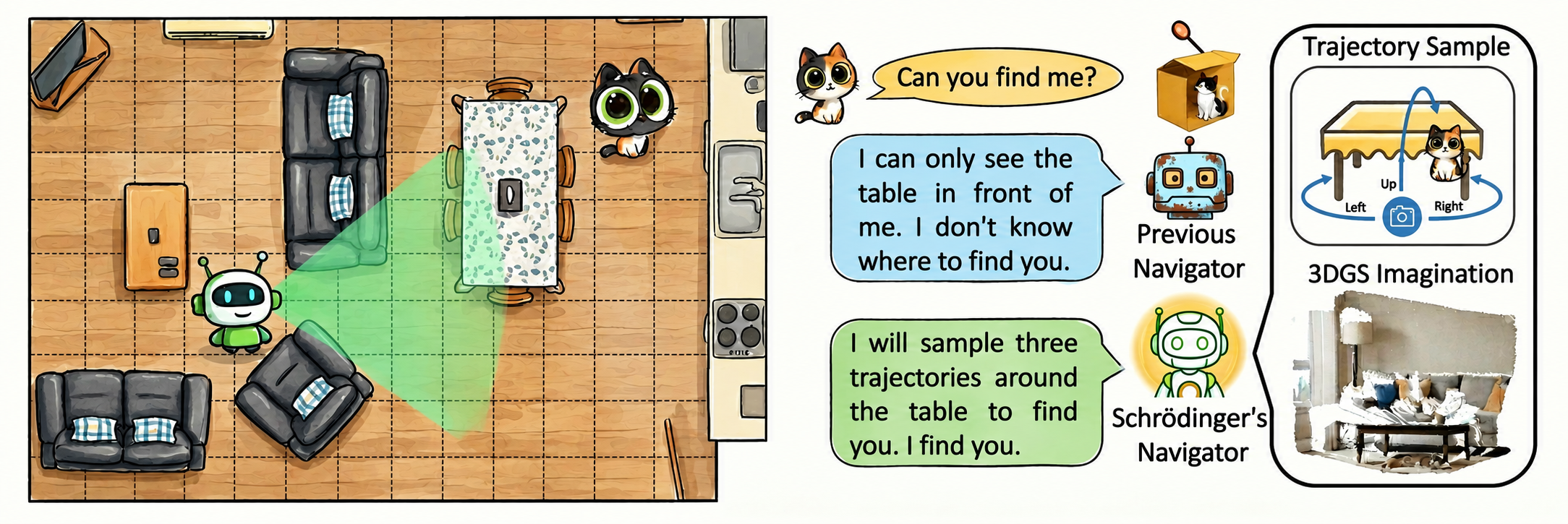} 

  \caption{Real-world zero-shot object navigation often fails when the target (e.g., a cat) is hidden behind occlusions and surrounded by hazardous space. Conventional navigation systems perceive only the immediate occluder and are unable to infer what exists beyond it. Our Schrödinger's Navigator addresses this challenge by modeling the unobserved regions as plausible futures.}

  \label{fig:teaser}

\end{figure}
\begin{abstract}
Zero-shot object navigation (ZSON) requires robots to find target objects in unseen environments without task-specific fine-tuning or pre-built maps, a key capability for general-purpose service robots. Yet methods that perform well in simulation often degrade in cluttered real-world scenes with severe occlusion and latent hazards, where large unseen regions make single-scene inference brittle and unsafe. We propose \textbf{Schrödinger's Navigator}, a belief-aware framework that reasons at inference time over multiple \emph{trajectory-conditioned imagined 3D futures}. Given candidate paths, a trajectory-conditioned 3D world model predicts hypothetical observations and maintains a superposition of plausible scene realizations rather than committing to one map. An adaptive occluder-aware sampler directs imagination to uncertainty-critical regions, while a \emph{Future-Aware Value Map} (FAVM) aggregates imagined futures for robust, proactive action selection. Experiments in simulation and on a physical Go2 quadruped show that Schrödinger's Navigator outperforms strong ZSON baselines, improving hidden-target discovery and risk-aware waypoint selection in occlusion-heavy navigation scenarios. These results highlight imagined 3D futures as a promising foundation for a modular, scalable, and generalizable framework for zero-shot navigation in uncertain real-world environments.
\end{abstract}


\section{Introduction}
\label{sec:intro}
Object navigation is a fundamental capability for mobile robots operating in human-centered environments \cite{chaplot2020object,anderson2018evaluation,cao2025cognav}. Practical service robots must find requested objects in previously unseen scenes without pre-built maps \cite{chaplot2020learning,chen2019behavioral,kwon2023renderable} or costly scene-specific retraining \cite{zhu2017target,ramrakhya2023pirlnav}. Zero-shot object navigation (ZSON) formalizes this setting by requiring a robot to locate a specified object in a novel environment without task-specific fine-tuning \cite{majumdar2022zson,gadre2023cows}. 

Despite strong results in simulation, current ZSON systems often degrade in cluttered real-world scenes with severe occlusions and latent hazards \cite{zhao2022zero,gadre2023cows,yokoyama2024vlfm,cai2024bridging}. We define a latent hazard as an initially unobserved, unannotated region that may cause collision, unsafe traversal, or task failure if traversed~\cite{zhang2021safe,firoozi2024oa,axelrod2018provably}. In these scenes, targets and risks may only become visible from revealing viewpoints, as illustrated by the occluded cat in \autoref{fig:teaser}. Yet many existing methods act on a single inferred scene interpretation, making them prone to missing hidden targets or choosing unsafe paths through uncertain regions \cite{racaniere2017imagination,ramakrishnan2020occupancy}.

To address this limitation, we draw inspiration from Schrödinger's thought experiment on uncertainty and propose \textbf{Schrödinger's Navigator}, a principled navigation framework that treats unobserved space as a superposition of plausible future worlds and reasons over them before committing to an action, as illustrated in \autoref{fig:teaser}. Rather than completing the scene once, the agent reasons at inference time over multiple \emph{trajectory-conditioned imagined 3D futures}. This enables the robot to evaluate how the world may appear along different candidate paths and plan not only from what is observed, but also from what could be revealed behind occlusions.

At the core of our framework is a trajectory-conditioned 3D world model~\cite{li2025flashworld}, which differs from prior world-model navigation methods~\cite{bar2025navigation,wang2025dreamnav,huang2025vistav2} in how imagined futures are generated and used. Given egocentric visual observations and sampled trajectories, the model predicts hypothetical 3D observations along each path, maintaining multiple plausible realizations of unobserved space. To make imagination efficient and decision-relevant, we introduce an \textit{adaptive, occluder-aware trajectory sampling} strategy that uses local geometry and uncertainty to focus predictions on regions where occlusion is most likely to affect navigation, avoiding costly dense rollouts over the full scene.
The imagined 3D predictions are spatially aligned and fused into a \emph{Future-Aware Value Map} (FAVM). Unlike value maps built only from observed states or short-horizon predictions, FAVM aggregates evidence across candidate futures and assigns value to currently unobserved regions according to their expected utility. The policy can reduce uncertainty, expose likely target locations, and avoid hazardous regions without requiring dense global reconstruction or extra exploratory detours.

We evaluate Schrödinger's Navigator in Habitat and on a physical Go2 quadruped across ObjectNav and occlusion-heavy scenarios. Results show that future-aware reasoning preserves competitive ObjectNav performance while improving behavior when targets or risks are initially hidden. By reasoning over imagined 3D futures at inference time, our method helps reveal hidden objects, avoid uncertain blind spots, and generalize to zero-shot navigation under real-world uncertainty.

In summary, our contributions are as follows:
\begin{itemize}
    \item We identify a fundamental limitation of existing zero-shot object navigation methods under severe occlusion and latent hazards: their reliance on single-scene inference without explicit reasoning over uncertainty in unobserved space.
    \item We propose \textbf{Schrödinger's Navigator}, a belief-aware navigation framework that reasons over multiple \emph{trajectory-conditioned imagined 3D futures}. The framework combines a trajectory-conditioned 3D world model, adaptive occluder-aware trajectory sampling, and a \emph{Future-Aware Value Map} (FAVM) for proactive uncertainty-aware planning.
    \item We demonstrate in both Habitat simulation and real-world Go2 deployment that explicit reasoning over imagined 3D futures improves robustness in occlusion-heavy object search, earlier hidden-target discovery, and risk-aware waypoint selection.
\end{itemize}
\begin{figure*}[t]  
    \centering
    \includegraphics[width=0.93\textwidth]{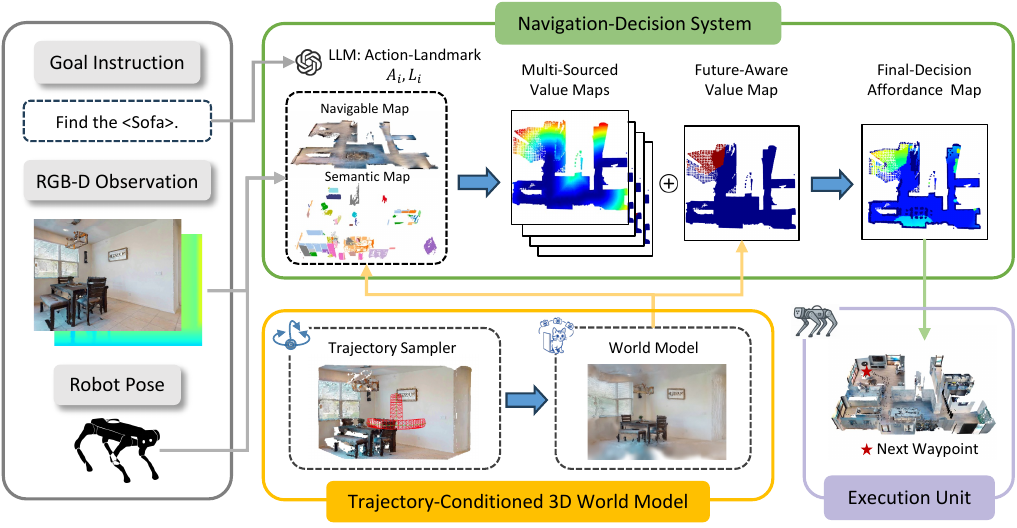} 
    \caption{Overview of the Schrödinger’s Navigator. Given the navigation context, the system adaptively samples candidate trajectories around occluded and uncertain regions. A trajectory-conditioned 3D world model then imagines future 3DGS observations along each path to expose hidden structures, target cues, and risks. These imagined futures are fused with current observations to construct a Future-Aware Value Map, aggregating trajectory-level utility and risk for robust waypoint selection. Finally, the selected navigation target is executed through the corresponding local controller.
    \label{fig:pipeline} }
\end{figure*}

\section{Related Work}
\label{sec:relatedwork}

\noindent \textbf{Object Navigation} (ObjectNav)~\cite{gupta2017cognitive} studies embodied agents that locate target objects in unseen environments. \emph{Task-trained} approaches based on reinforcement or imitation learning \cite{ramrakhya2023pirlnav,chaplot2020object,mayo2021visual,ramrakhya2022habitat} require large-scale training and often generalize poorly to cluttered real-world scenes, making sim-to-real deployment challenging. 
\emph{Zero-shot} approaches instead leverage pretrained vision-language models \cite{li2023blip,majumdar2022zson,gadre2023cows} or large language models \cite{yang2023dawn,zhou2023esc,shah2023lm} for reasoning and planning without task-specific training \cite{yokoyama2024vlfm,gadre2023cows,shah2023lm,zhou2026beliefmapnav}. These methods improve semantics-driven exploration through multimodal target embeddings \cite{majumdar2022zson,gadre2023cows}, vision-language frontier maps \cite{yokoyama2024vlfm}, instruction-based prompting \cite{long2024instructnav,zhou2023esc}, and adaptive fusion of semantic and geometric cues \cite{kuang2024openfmnav,huang2024gamap,chen2023not,zhang2024imagine,zhang2025apexnav}. 
However, zero-shot ObjectNav remains brittle in realistic, occlusion-heavy environments and under unknown risks \cite{gadre2023cows,yokoyama2024vlfm}, because value estimation is still dominated by visible frontiers and lacks explicit reasoning about hidden areas behind occluders, leading to myopic exploration.

\noindent \textbf{Imagination for Navigation} leverages generation models to simulate future observations and support planning \cite{koh2021pathdreamer,bar2025navigation,qin2025navigatediff,wang2025dreamnav}. Related world-model methods learn predictive dynamics and roll out trajectories in latent space \cite{liu2025x,wang2023dreamwalker}. Recent navigation world models further predict egocentric visual streams. NWM \cite{bar2025navigation} predicts future trajectories and views for path ranking, while NavigateDiff \cite{qin2025navigatediff} uses diffusion-based visual prediction as a zero-shot navigation assistant. 
Language-conditioned variants, such as VISTA and VISTAv2 \cite{perincherry2025visual,huang2025vista,huang2025vistav2}, generate landmark images or instruction-conditioned future views and use their alignment to guide action selection. Other works imagine unseen scenes or predict occupancy maps to complete unobserved spaces for exploration \cite{katyal2019uncertainty,sharma2023proxmap,shah2025foresightnav,zhao2025imaginenav,wang2026imaginenav++}. These works are mainly studied under visual prediction, or vision-and-language navigation protocols. Unlike prior works, Schrödinger's Navigator imagines future observations along multiple paths with a trajectory-conditioned 3D world model and constructs a FAVM, enabling explicit reasoning about occlusions, risks, and blind-spot targets for robust zero-shot navigation.

\section{Method}

We introduce \textbf{Schr\"odinger's Navigator}, a zero-shot object navigation framework that reduces occlusion-induced uncertainty by reasoning over multiple trajectory-conditioned imagined 3D futures. At each decision step, the robot receives the current navigation context, including the goal instruction, RGB-D observation, robot pose, and accumulated map. Instead of committing to a single inferred scene, our method samples a compact set of occluder-aware candidate trajectories, imagines future 3D scenes along them, and aggregates the resulting cues into a future-aware affordance map for waypoint selection.

\subsection{Trajectory-Conditioned 3D Future Imagination}
\paragraph{Adaptive occluder-aware trajectory sampling.}

\begin{wrapfigure}{r}{0.25\linewidth} 
    \centering
    \includegraphics[width=\linewidth]{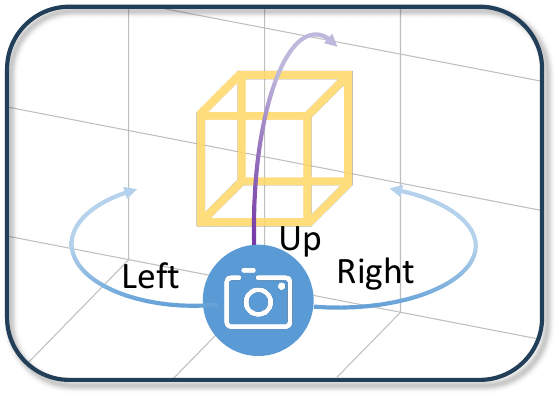} 
    \caption{Occluder-aware camera trajectory sampling.}
    \label{fig:traj}
\end{wrapfigure}

A single imagined path may miss targets or hazards hidden behind occlusions. To obtain complementary views with limited computation, we sample a small trajectory ensemble around the dominant occluder in the view. Specifically, we construct a local point cloud from the depth observation, remove ground points, and select the primary forward occluder according to projected occupancy, distance, and alignment with the robot's heading. Its boundary provides a local geometric anchor for trajectory generation.

As shown in \autoref{fig:traj}, we instantiate three trajectory types
\(\mathcal{V}=\{L,R,U\}\), corresponding to left-bypass, right-bypass, and over-the-top views. For each type \(v\in\mathcal{V}\), the sampler generates a camera trajectory
\(\tau_t^{(v)}=\mathcal{F}_{\rm samp}(v;B_t,q_t,d_c,N_{\rm pose})\),
where \(B_t\) is the estimated occluder boundary, \(q_t\) is the current robot pose, \(d_c\) is a nominal orbit-distance parameter, and \(N_{\rm pose}\) is the number of sampled camera poses. This compact trajectory set is designed to cover potentially visible regions on both sides of and above the occluder while avoiding dense sampling over the full action space. The three trajectories are generated from the same RGB-D observation to provide complementary virtual camera viewpoints, rather than independent world-state samples or robot-executed paths. Details of occluder boundary extraction, occluder selection, and trajectory parameterization are provided in Appendix~B.

\begin{figure}[!t]  
    \centering
    \includegraphics[width=0.93\textwidth]{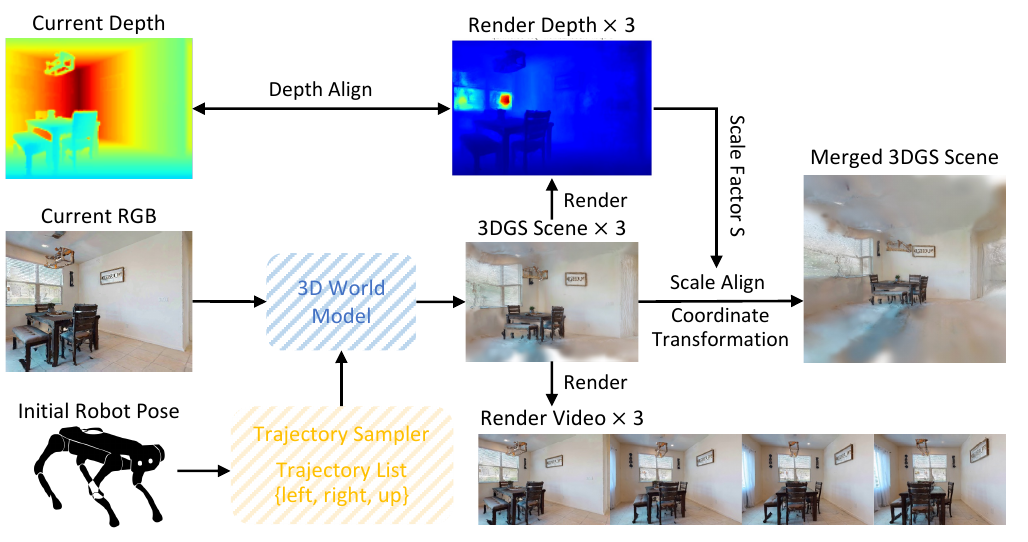} 
    \caption{Overview of trajectory-conditioned 3D imagination: Given the current observation and robot pose, an occluder-aware sampler generates candidate trajectories that condition a 3D world model to predict future 3DGS scenes. These scenes are depth-aligned to recover a consistent scale, transformed into the world frame, and fused into a unified 3DGS scene for future-aware planning.  \label{fig:3dgs} }
\end{figure}

\paragraph{Trajectory-conditioned 3D world model.}
As shown in \autoref{fig:3dgs}, given each sampled trajectory \(\tau_t^{(v)}\), we use a trajectory-conditioned 3D world model \(\Phi_\theta\) to generate a hypothetical future 3DGS:
\[
\hat{\mathcal{G}}_t^{(v)}
=
\Phi_\theta
\left(I_t, q_t, \tau_t^{(v)}\right),
\label{eq:world_model}
\]
where \(I_t\) denotes the current RGB observation. We adopt FlashWorld~\cite{li2025flashworld} as the backend because it can generate 3D-consistent scenes efficiently. Each generated scene represents a plausible realization of the currently unobserved region conditioned on one candidate path.

Since the generated 3DGS scene may not be metrically aligned with the robot's environment, we use the observed depth \(D_t\) to perform metric scale alignment before fusion. We first render a depth map \(\hat{D}_t^{(v)}\) from the generated scene at the current viewpoint and estimate a global scale factor using valid pixels:
\[
s_t^{(v)}
=
\operatorname*{median}_{p\in\Omega_t^{(v)}}
\frac{D_t(p)}{\hat{D}_t^{(v)}(p)} ,
\label{eq:scale_alignment}
\]
where \(\Omega_t^{(v)}\) is the set of pixels with valid observed and rendered depths. The generated scene is then scaled and transformed into the world frame as
\(\tilde{\mathcal{G}}_t^{(v)}=\mathcal{A}(\hat{\mathcal{G}}_t^{(v)},s_t^{(v)},q_t)\),
where \(\mathcal{A}(\cdot)\) denotes scale alignment and coordinate transformation. The aligned scenes are fused into a unified future scene
\(\mathcal{G}_t^{\rm img}=\bigcup_{v\in\mathcal{V}}\tilde{\mathcal{G}}_t^{(v)}\).
We further transfer semantic labels from 2D segmentation to visible Gaussian primitives using depth-consistent projection and multi-view voting. Implementation details of 3DGS alignment, semantic label transfer, and scene fusion are provided in Appendix~C.



\subsection{Future-Aware Navigation Decision}
\label{sec:decision}

\paragraph{Future-Aware Value Map.}
\label{sec:favm}
The imagined scene \(\mathcal{G}_t^{\rm img}\) provides cues about free space, possible target locations, and risky regions beyond the current observation. We convert these cues into a \emph{Future-Aware Value Map} (FAVM), denoted by \(m_t^{\rm FA}\), over candidate navigation positions \(\mathcal{P}_t=\{\mu_g\mid g\in\mathcal{G}_{\rm nav}\}\), where \(\mu_g\) is the center of a navigable Gaussian:
\[
m_t^{\rm FA}(p)
=
\mathcal N\!\left(
\alpha_{\rm sem} S_t(p)
+
\alpha_{\rm exp} E_t(p)
-
\alpha_{\rm risk} R_t(p)
\right),
\qquad p\in\mathcal{P}_t,
\label{eq:favm}
\]
where \(S_t(p)\) measures semantic relevance to the target, \(E_t(p)\) estimates the local coverage of newly imagined free space, and \(R_t(p)\) penalizes candidate paths passing near geometrically supported occupied regions in the imagined scenes. The weights \(\alpha_{\rm sem}\), \(\alpha_{\rm exp}\), and \(\alpha_{\rm risk}\) balance target seeking, exploration, and safety.

Specifically, $S_t(p)$ uses exponentially decayed nearest distances to observed and imagined target-labeled Gaussians, where imagined target cues are discounted by a factor $\lambda_{\mathrm{img}}<1$ to avoid over-trusting generated content. 
$E_t(p)$ estimates the amount of previously unobserved but likely navigable space that can be estimated around $p$, computed from imagined free-space primitives not yet covered by the observed map. 
$R_t(p)$ is a soft geometric risk penalty computed from filtered and spatially deduplicated imagined occupied voxels according to their distances from the candidate path. 
All three component scores are normalized to $[0,1]$ before fusion, and the combined FAVM score is normalized once more. Implementation details, including kernel radius, semantic voting, and normalization, are provided in Appendix~D.

\paragraph{Fusion with observation-based cues.}
FAVM provides foresight, but navigation must remain grounded in current observations. We combine FAVM with an observation-based multi-sourced value map
\(m_t=m_t^{\rm act}+m_t^{\rm sem}+m_t^{\rm traj}+m_t^{\rm heu}\),
which aggregates action guidance, semantic landmark cues, trajectory constraints, and heuristic navigation priors.

The final affordance map is computed as
\[
m_t^{\rm aff}(p)
=
\beta \, \bar{m}_t(p)
+
(1-\beta) \, m_t^{\rm FA}(p),
\qquad
p\in\mathcal{P}_t,
\label{eq:affordance_fusion}
\]
where \(\bar{m}_t\) is the normalized observation-based value map and \(\beta\in[0,1]\) controls the balance between grounded evidence and imagined future guidance. The navigation target is selected as
\(p_t^\star=\arg\max_{p\in\mathcal{P}_t}m_t^{\rm aff}(p)\),
with walkability constraints enforced during selection. Detailed definitions of FAVM scores and observation-based component maps are provided in Appendix~D.

\paragraph{Execution.}
Following an instruction-driven hierarchical planner, the goal instruction is parsed into high-level action-landmark guidance, which contributes to the observation-based value map. The selected navigation target \(p_t^\star\) is passed to the local planner to compute executable waypoints for robot control. To reduce latency, the world model is invoked periodically rather than at every navigation step.

   

\section{Experiments}
\label{sec:experiments}

We evaluate Schrödinger's Navigator in both simulation and real-world settings to assess its navigation performance, component effectiveness, and practical deployability. Our experiments first compare against strong ZSON baselines on HM3D, then ablate the key design choices behind future imagination and FAVM-based planning, and finally validate the system on a physical Go2 quadruped in indoor environments. More qualitative results and failure cases are shown in Appendix~F.

\subsection{Experimental Setup}
\label{sec:exp_setup}

\paragraph{Baselines and benchmark.} We compare our approach against a broad set of representative baselines, including 
task-trained approaches (ZSON~\cite{majumdar2022zson}, 
PixNav~\cite{cai2024bridging}, SPNet~\cite{zhao2023semantic}, 
SGM~\cite{zhang2024imagine}), 
zero-shot semantic or map-based explorers (CoW~\cite{gadre2023cows}, 
ESC~\cite{zhou2023esc}, L3MVN~\cite{yu2023l3mvn}, 
TriHelper~\cite{zhang2024trihelper}, VoroNav~\cite{wu2024voronav}, 
GAMap~\cite{huang2024gamap}, BeliefMapNav~\cite{zhou2026beliefmapnav}), 
imagination-based VLM navigators (ImagineNav~\cite{zhao2025imaginenav}, 
ImagineNav++~\cite{wang2026imaginenav++}), 
and LLM/VLM-based planners (VLFM~\cite{yokoyama2024vlfm}, 
InstructNav~\cite{long2024instructnav}).
We conduct large-scale simulation evaluation on HM3D~\cite{ramakrishnan2021hm3d} 
within Habitat~\cite{puig2023habitat3}. 
Following prior work, we report Success Rate (SR), Success weighted by Path Length (SPL), 
and Distance to Goal (DTG). 
Detailed descriptions of baselines are provided in Appendix~E.

\paragraph{Implementation.}
We use Qwen3.6-35B-A3B~\cite{qwen36_35b_a3b} for instruction parsing and navigation-direction judgment, FlashWorld~\cite{li2025flashworld} for trajectory-conditioned 3D imagination, and GLEE~\cite{wu2024general} for semantic segmentation. In HM3D simulation, FlashWorld is triggered every 50 observation updates, rather than at every step. Real-world tests use a Unitree Go2 with an RGB-D camera in Office, Classroom, and Common Room scenes, covering static-object search and occlusion-heavy unknown-risk cases. Hyperparameters, runtime, compute, and deployment details are provided in Appendices~A and~G.


\subsection{Simulation Results}
\label{sec:sim_eval}


\begin{table}[t]
    \centering
    \caption{\textbf{Quantitative Comparison on Simulation Results.} Cell background colors indicate the method is the {\colorbox{red!50}{best}}, {\colorbox{orange!50}{second best}}, or {\colorbox{yellow!50}{third best}} on this metric.}
    {
        \begin{tabular}{l c  ccc}
        \toprule
        \multirow{2}{*}{\textbf{Method}} & \multirow{2}{*}{\textbf{Training Free}}
        & \multicolumn{3}{c}{\textbf{HM3D}} \\
        & & SR$\uparrow$ & SPL$\uparrow$ & DTG$\downarrow$ \\
        \midrule
        ZSON~\cite{majumdar2022zson}        & \xmark      & 0.255 & 0.126 & --   \\
        PixNav~\cite{cai2024bridging}      & \xmark      & 0.379 & 0.205 & --   \\
        SPNet~\cite{zhao2023semantic}      & \xmark      & 0.312 & 0.101 & --   \\
        SGM~\cite{zhang2024imagine}        & \xmark      & 0.602 & \cellcolor{yellow!50}0.308 & -- \\
        \midrule
        ESC~\cite{zhou2023esc}        & \cmark  & 0.392 & 0.223 & --   \\
        VLFM~\cite{yokoyama2024vlfm}       & \cmark  & 0.525 & 0.304 & --   \\
        VoroNav~\cite{wu2024voronav}    & \cmark  & 0.420 & 0.260 & --   \\
        L3MVN~\cite{yu2023l3mvn}      & \cmark  & 0.504 & 0.231 & 4.43 \\
        TriHelper~\cite{zhang2024trihelper}  & \cmark  & 0.565 & 0.253 & \cellcolor{yellow!50}3.87 \\
        GAMap~\cite{huang2024gamap}      & \cmark  & 0.531 & 0.260 & --   \\
        InstructNav~\cite{long2024instructnav} & \cmark  & 0.510 & 0.187 & \cellcolor{orange!50}2.89 \\
        ImagineNav~\cite{zhao2025imaginenav}    & \cmark  & 0.530 & 0.238 & -- \\
        ImagineNav++~\cite{wang2026imaginenav++}    & \cmark  & \cellcolor{orange!50}0.625 & \cellcolor{red!50}0.328 & -- \\
        BeliefMapNav~\cite{zhou2026beliefmapnav}    & \cmark  & \cellcolor{yellow!50}0.614 & 0.306 & -- \\
        \midrule
        \textbf{Ours} & \cmark  & \cellcolor{red!50}0.682 & \cellcolor{orange!50}0.311 & \cellcolor{red!50}1.84 \\
        \bottomrule
        \end{tabular}
    }
    \label{tab:sim_result}
\end{table}

\subsubsection{Quantitative Results}
Table~\ref{tab:sim_result} presents the quantitative results on HM3D.
Our method achieves the highest SR of 0.682, outperforming the strongest baseline, ImagineNav++ (0.625), by 5.7 percentage points.
It also attains the lowest reported DTG of 1.84 m, compared with 2.89 m for InstructNav.
Meanwhile, our method achieves the second-highest SPL of 0.311, remaining competitive in path efficiency.
These results demonstrate the effectiveness of future-aware navigation in improving target discovery and goal proximity while maintaining competitive path efficiency.
Further analysis across different occlusion severity levels is provided in Appendix~\ref{app:baselines} (Table~\ref{tab:occlusion_severity}).

\subsubsection{Ablation Analysis}
\label{sec:ablation}

To assess the key design choices of Schr\"odinger's Navigator, we ablate the trajectory ensemble size $N$, the fusion coefficient $\beta$, and the FAVM scoring components. All variants change one factor from the full model, except for the no-imagination baseline.

\begin{table}[t]
\centering
\caption{
Ablation and runtime analysis of Schr\"odinger's Navigator on HM3D.
We report success rate (SR), average number of imagination triggers per episode,
per-trigger imagination latency, episode-level waiting ratio, and waypoint change rate.
}
\label{tab:ablation}
\begin{tabular}{@{}lccccc@{}}
\toprule
Variant 
& SR $\uparrow$ 
& \makecell{Trig.\\ / Ep. $\downarrow$}
& \makecell{Lat.\\ / Trig. (s) $\downarrow$}
& \makecell{Wait\\ Ratio (\%) $\downarrow$}
& \makecell{Waypoint\\ Change (\%)} \\
\midrule
No Imagination 
& 0.553 & 0.00 & -- & 0.0 & -- \\
\midrule
Single Traj. $(N=1)$ 
& 0.594 & 4.31 & 3.0 & 7.2 & 28.4 \\

Dense Sampling $(N=5)$ 
& 0.622 & 4.18 & 15.0 & 30.8 & 51.6 \\
\midrule
Imagination-heavy $(\beta=0.2)$ 
& 0.497 & 4.36 & 9.0 & 22.9 & 67.2 \\

Observation-heavy $(\beta=0.8)$ 
& 0.595 & 4.28 & 9.0 & 19.9 & 35.3 \\
\midrule
FAVM w/o Exploration $E(p)$ 
& 0.629 & 4.20 & 9.0 & 19.8 & 40.8 \\

FAVM w/o Semantic $S(p)$ 
& 0.605 & 4.27 & 9.0 & 20.5 & 38.4 \\

FAVM w/o Risk $R(p)$ 
& 0.634 & 4.19 & 9.0 & 19.3 & 42.1 \\
\midrule
Ours $(N=3,\beta=0.6)$ 
& \textbf{0.682} & 4.24 & 9.0 & 20.1 & 46.8 \\
\bottomrule
\end{tabular}
\end{table}

The results in \autoref{tab:ablation} show that future imagination is effective only when properly sampled and grounded. 
Our full model improves SR from 0.553 to 0.682 over the no-imagination baseline, a gain of 12.9 percentage points. 
However, more trajectories do not necessarily yield better performance: $N=1$ provides limited coverage, while $N=5$ incurs higher latency and still underperforms our $N=3$ design, demonstrating the coverage--cost advantage of tri-trajectory sampling.

The fusion and component ablations further show that imagined guidance must be calibrated by real observations. 
The imagination-heavy setting ($\beta=0.2$) achieves an SR of 0.497, while the observation-heavy setting ($\beta=0.8$) reaches 0.595. Both underperform the balanced setting ($\beta=0.6$), highlighting the importance of balancing imagined future guidance with current observations. 
Removing any FAVM component reduces SR, with the largest drop caused by removing the semantic score $S(p)$.

The runtime results indicate that 3D future imagination introduces sparse rather than per-step overhead. 
Our full model triggers imagination 4.24 times per episode, with 9.0 seconds per trigger and a 20.1\% waiting-time ratio. 
Dense sampling increases the waiting-time ratio to 30.8\% without improving SR, showing that additional futures introduce extra cost without proportional benefit. 
The 46.8\% waypoint change rate further indicates that imagined futures actively influence waypoint selection beyond the observation-only map.

\subsection{Real-World Deployment}
\label{sec:real_eval}

Our real-world experiments provide a controlled evaluation against an InstructNav-style baseline, complementing the broader comparisons with navigation methods conducted on HM3D.







We evaluate adaptability under two task settings: 
(1) \textbf{Static Object Search}, where the agent explores standard scenes to locate objects; and 
(2) \textbf{Unknown-Risk Scenarios}, where severe static occlusions are not visible from afar, testing the agent's ability to anticipate uncertainty and avoid blind spots.

\begin{figure*}[t]  
    \centering
    \includegraphics[width=0.95\textwidth]{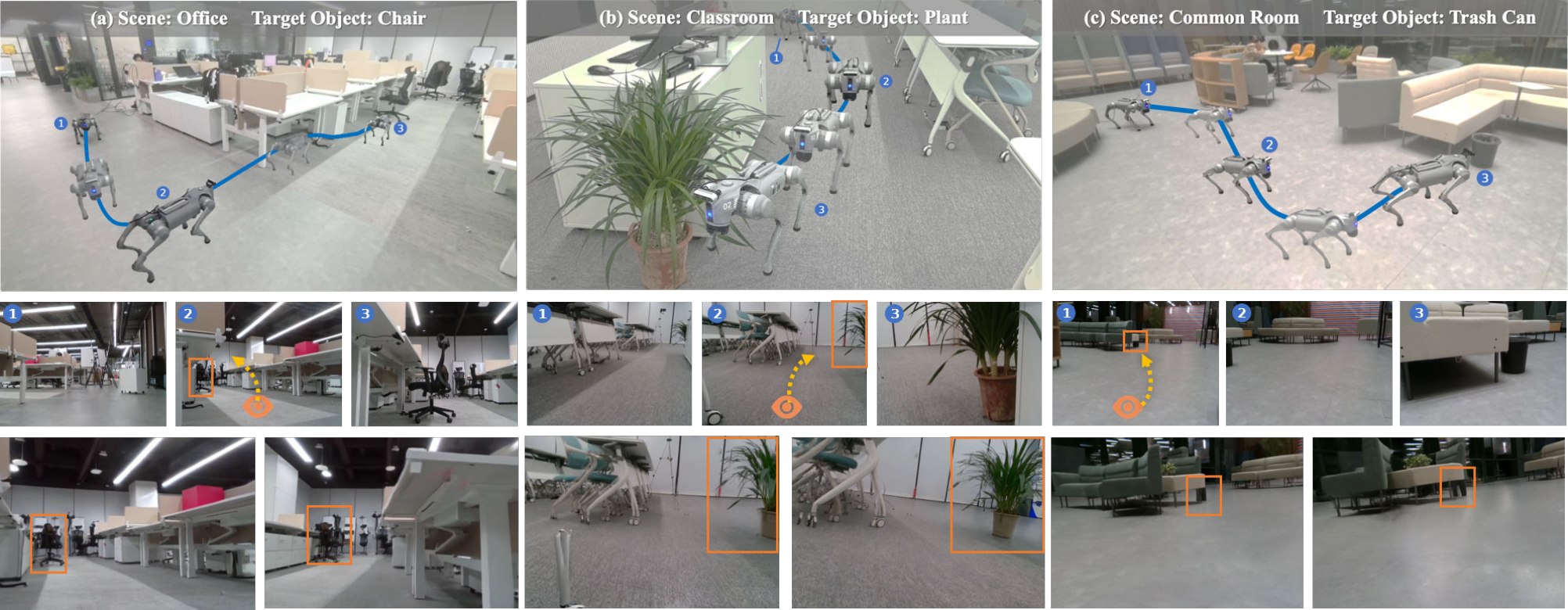} 
    \caption{Real-world object-navigation demonstrations of Schrödinger’s Navigator across three indoor scenarios: Office-Chair, Classroom-Plant, and Common Room-Trash Can. Each example shows the executed trajectory, egocentric observations with predicted target directions and target boxes, and world-model imagined views that reveal the target appearance.
        \label{experiments} }
\end{figure*}

\begin{table}[t] \small
    \centering
    \caption{Real-world quantitative comparison against the InstructNav baseline. We report success counts (Success / Total Trials) across three indoor environments.}
    \label{Comparison with baseline method}
    {
        \begin{tabular}{ccccc}
            \toprule
            \textbf{Scene} & \textbf{Office} & \textbf{Classroom} & \textbf{Common Room} & \textbf{All} \\
            \midrule
            \multicolumn{5}{c}{\textbf{Search for static objects}} \\
            \midrule
            InstructNav & 7/10 & 7/10 & \textbf{8/10} & 22/30 \\
            Ours & \textbf{8/10} & \textbf{8/10} & 7/10 & \textbf{23/30} \\
            \midrule
            \multicolumn{5}{c}{\textbf{Scenarios with Unknown Risks}} \\
            \midrule
            InstructNav & 4/10 & 3/10 & 4/10 & 11/30 \\
            Ours & \textbf{5/10} & \textbf{6/10} & \textbf{6/10} & \textbf{17/30} \\
            \bottomrule
        \end{tabular}
    } 
\end{table}

\subsubsection{Performance Analysis}
As shown in \autoref{Comparison with baseline method}, we report success counts over 30 trials per setting, with 10 trials in each of the three environments.

In standard static object search, Schrödinger's Navigator performs comparably to InstructNav, achieving a slightly higher success rate ($23/30$ vs. $22/30$), indicating that future-aware reasoning does not compromise regular object-search performance. In scenarios with unknown risks, our method shows a clearer advantage, succeeding in $17/30$ trials compared with InstructNav's $11/30$. This suggests that imagined 3D futures help the robot better handle occlusions, ambiguous observations, and potential blind spots.

Overall, these real-world results show that our method maintains strong static-search performance while improving robustness in risk-aware navigation. Qualitative examples are provided in \autoref{experiments}.

\section{Conclusion}

Schrödinger's Navigator addresses ZSON limitations in occluded environments by leveraging a 3D trajectory-conditioned world model to imagine and fuse plausible 3D futures into a unified scene. This training-free framework constructs a single affordance map encoding semantics, exploration, and risk. Real-world experiments on a Unitree Go2 demonstrate its robustness, achieving comparable performance in static tasks while outperforming the baseline in occlusion-heavy scenarios.

\section{Limitations}
Schrödinger's Navigator uses imagined 3D futures as plausible hypotheses rather than exact reconstructions of hidden space. While FAVM discounts generated cues and grounds decisions in current observations, inaccurate generation, semantic transfer errors, or imperfect depth alignment may still bias waypoint selection when imagined evidence is over-weighted. Our method performs risk-aware waypoint selection but does not provide formal safety guarantees. Future work could improve reliability through uncertainty-calibrated imagination, stronger multi-view consistency, and active verification of uncertain regions. In addition, the current tri-trajectory sampler focuses on the dominant forward occluder for efficiency, but more complex layouts with multiple occluders or non-frontal blind spots may require adaptive trajectory budgeting. Broader evaluation across more diverse environments, dynamic obstacles, sensors, and robot platforms, as well as integration with certified local planners, would further strengthen real-world deployment.



\clearpage
\acknowledgments{This work was supported in part by the Science and Technology Commission of Shanghai Municipality (No. 24511103100) and the Xiaomi Young Scholar Program.}


\bibliography{references}  

@inproceedings{gupta2017cognitive,
  title={Cognitive mapping and planning for visual navigation},
  author={Gupta, Saurabh and Davidson, James and Levine, Sergey and Sukthankar, Rahul and Malik, Jitendra},
  booktitle={Proceedings of the IEEE conference on computer vision and pattern recognition},
  pages={2616--2625},
  year={2017}
}

@article{anderson2018evaluation,
  title={On evaluation of embodied navigation agents},
  author={Anderson, Peter and Chang, Angel and Chaplot, Devendra Singh and Dosovitskiy, Alexey and Gupta, Saurabh and Koltun, Vladlen and Kosecka, Jana and Malik, Jitendra and Mottaghi, Roozbeh and Savva, Manolis and others},
  journal={arXiv preprint arXiv:1807.06757},
  year={2018}
}

@article{chaplot2020object,
  title={Object goal navigation using goal-oriented semantic exploration},
  author={Chaplot, Devendra Singh and Gandhi, Dhiraj Prakashchand and Gupta, Abhinav and Salakhutdinov, Russ R},
  journal={Advances in Neural Information Processing Systems},
  volume={33},
  pages={4247--4258},
  year={2020}
}

@article{zhao2022zero,
  title={Zero-shot object goal visual navigation},
  author={Zhao, Qianfan and Zhang, Lu and He, Bin and Qiao, Hong and Liu, Zhiyong},
  journal={arXiv preprint arXiv:2206.07423},
  year={2022}
}

@inproceedings{cai2024bridging,
  title={Bridging zero-shot object navigation and foundation models through pixel-guided navigation skill},
  author={Cai, Wenzhe and Huang, Siyuan and Cheng, Guangran and Long, Yuxing and Gao, Peng and Sun, Changyin and Dong, Hao},
  booktitle={2024 IEEE International Conference on Robotics and Automation (ICRA)},
  pages={5228--5234},
  year={2024},
  organization={IEEE}
}

@inproceedings{gadre2023cows,
  title={Cows on pasture: Baselines and benchmarks for language-driven zero-shot object navigation},
  author={Gadre, Samir Yitzhak and Wortsman, Mitchell and Ilharco, Gabriel and Schmidt, Ludwig and Song, Shuran},
  booktitle={Proceedings of the IEEE/CVF Conference on Computer Vision and Pattern Recognition},
  pages={23171--23181},
  year={2023}
}

@inproceedings{yokoyama2024vlfm,
  title={Vlfm: Vision-language frontier maps for zero-shot semantic navigation},
  author={Yokoyama, Naoki and Ha, Sehoon and Batra, Dhruv and Wang, Jiuguang and Bucher, Bernadette},
  booktitle={2024 IEEE International Conference on Robotics and Automation (ICRA)},
  pages={42--48},
  year={2024},
  organization={IEEE}
}

@article{chaplot2020learning,
  title={Learning to explore using active neural slam},
  author={Chaplot, Devendra Singh and Gandhi, Dhiraj and Gupta, Saurabh and Gupta, Abhinav and Salakhutdinov, Ruslan},
  journal={arXiv preprint arXiv:2004.05155},
  year={2020}
}

@inproceedings{ramakrishnan2020occupancy,
  title={Occupancy anticipation for efficient exploration and navigation},
  author={Ramakrishnan, Santhosh K and Al-Halah, Ziad and Grauman, Kristen},
  booktitle={European conference on computer vision},
  pages={400--418},
  year={2020},
  organization={Springer}
}

@article{majumdar2022zson,
  title={Zson: Zero-shot object-goal navigation using multimodal goal embeddings},
  author={Majumdar, Arjun and Aggarwal, Gunjan and Devnani, Bhavika and Hoffman, Judy and Batra, Dhruv},
  journal={Advances in Neural Information Processing Systems},
  volume={35},
  pages={32340--32352},
  year={2022}
}

@article{racaniere2017imagination,
  title={Imagination-augmented agents for deep reinforcement learning},
  author={Racani{\`e}re, S{\'e}bastien and Weber, Th{\'e}ophane and Reichert, David and Buesing, Lars and Guez, Arthur and Jimenez Rezende, Danilo and Puigdom{\`e}nech Badia, Adri{\`a} and Vinyals, Oriol and Heess, Nicolas and Li, Yujia and others},
  journal={Advances in neural information processing systems},
  volume={30},
  year={2017}
}

@inproceedings{bar2025navigation,
  title={Navigation world models},
  author={Bar, Amir and Zhou, Gaoyue and Tran, Danny and Darrell, Trevor and LeCun, Yann},
  booktitle={Proceedings of the Computer Vision and Pattern Recognition Conference},
  pages={15791--15801},
  year={2025}
}

@inproceedings{ramrakhya2023pirlnav,
  title={Pirlnav: Pretraining with imitation and rl finetuning for objectnav},
  author={Ramrakhya, Ram and Batra, Dhruv and Wijmans, Erik and Das, Abhishek},
  booktitle={Proceedings of the IEEE/CVF Conference on Computer Vision and Pattern Recognition},
  pages={17896--17906},
  year={2023}
}

@inproceedings{mayo2021visual,
  title={Visual navigation with spatial attention},
  author={Mayo, Bar and Hazan, Tamir and Tal, Ayellet},
  booktitle={Proceedings of the IEEE/CVF conference on computer vision and pattern recognition},
  pages={16898--16907},
  year={2021}
}

@inproceedings{ramrakhya2022habitat,
  title={Habitat-web: Learning embodied object-search strategies from human demonstrations at scale},
  author={Ramrakhya, Ram and Undersander, Eric and Batra, Dhruv and Das, Abhishek},
  booktitle={Proceedings of the IEEE/CVF conference on computer vision and pattern recognition},
  pages={5173--5183},
  year={2022}
}

@inproceedings{shah2023lm,
  title={Lm-nav: Robotic navigation with large pre-trained models of language, vision, and action},
  author={Shah, Dhruv and Osi{\'n}ski, B{\l}a{\.z}ej and Levine, Sergey and others},
  booktitle={Conference on robot learning},
  pages={492--504},
  year={2023},
  organization={PMLR}
}

@article{long2024instructnav,
  title={Instructnav: Zero-shot system for generic instruction navigation in unexplored environment},
  author={Long, Yuxing and Cai, Wenzhe and Wang, Hongcheng and Zhan, Guanqi and Dong, Hao},
  journal={arXiv preprint arXiv:2406.04882},
  year={2024}
}

@article{kuang2024openfmnav,
  title={Openfmnav: Towards open-set zero-shot object navigation via vision-language foundation models},
  author={Kuang, Yuxuan and Lin, Hai and Jiang, Meng},
  journal={arXiv preprint arXiv:2402.10670},
  year={2024}
}

@article{qin2025navigatediff,
  title={Navigatediff: Visual predictors are zero-shot navigation assistants},
  author={Qin, Yiran and Sun, Ao and Hong, Yuze and Wang, Benyou and Zhang, Ruimao},
  journal={arXiv preprint arXiv:2502.13894},
  year={2025}
}

@inproceedings{perincherry2025visual,
  title={Do Visual Imaginations Improve Vision-and-Language Navigation Agents?},
  author={Perincherry, Akhil and Krantz, Jacob and Lee, Stefan},
  booktitle={Proceedings of the Computer Vision and Pattern Recognition Conference},
  pages={3846--3855},
  year={2025}
}

@article{huang2025vista,
  title={VISTA: Generative Visual Imagination for Vision-and-Language Navigation},
  author={Huang, Yanjia and Wu, Mingyang and Li, Renjie and Tu, Zhengzhong},
  journal={arXiv preprint arXiv:2505.07868},
  year={2025}
}

@inproceedings{shah2025foresightnav,
  title={Foresightnav: Learning scene imagination for efficient exploration},
  author={Shah, Hardik and Xing, Jiaxu and Messikommer, Nico and Sun, Boyang and Pollefeys, Marc and Scaramuzza, Davide},
  booktitle={Proceedings of the Computer Vision and Pattern Recognition Conference},
  pages={5236--5245},
  year={2025}
}

@inproceedings{sharma2023proxmap,
  title={Proxmap: Proximal occupancy map prediction for efficient indoor robot navigation},
  author={Sharma, Vishnu D and Chen, Jingxi and Tokekar, Pratap},
  booktitle={2023 IEEE/RSJ International Conference on Intelligent Robots and Systems (IROS)},
  pages={7135--7140},
  year={2023},
  organization={IEEE}
}

@article{li2025flashworld,
  title={FlashWorld: High-quality 3D Scene Generation within Seconds},
  author={Li, Xinyang and Wang, Tengfei and Gu, Zixiao and Zhang, Shengchuan and Guo, Chunchao and Cao, Liujuan},
  journal={arXiv preprint arXiv:2510.13678},
  year={2025}
}

@inproceedings{wu2024general,
  title={General object foundation model for images and videos at scale},
  author={Wu, Junfeng and Jiang, Yi and Liu, Qihao and Yuan, Zehuan and Bai, Xiang and Bai, Song},
  booktitle={Proceedings of the IEEE/CVF Conference on Computer Vision and Pattern Recognition},
  pages={3783--3795},
  year={2024}
}

@inproceedings{liu2025x,
  title={X-mobility: End-to-end generalizable navigation via world modeling},
  author={Liu, Wei and Zhao, Huihua and Li, Chenran and Biswas, Joydeep and Okal, Billy and Goyal, Pulkit and Chang, Yan and Pouya, Soha},
  booktitle={2025 IEEE International Conference on Robotics and Automation (ICRA)},
  pages={7569--7576},
  year={2025},
  organization={IEEE}
}

@inproceedings{wang2023dreamwalker,
  title={Dreamwalker: Mental planning for continuous vision-language navigation},
  author={Wang, Hanqing and Liang, Wei and Van Gool, Luc and Wang, Wenguan},
  booktitle={Proceedings of the IEEE/CVF international conference on computer vision},
  pages={10873--10883},
  year={2023}
}

@inproceedings{koh2021pathdreamer,
  title={Pathdreamer: A world model for indoor navigation},
  author={Koh, Jing Yu and Lee, Honglak and Yang, Yinfei and Baldridge, Jason and Anderson, Peter},
  booktitle={Proceedings of the IEEE/CVF International Conference on Computer Vision},
  pages={14738--14748},
  year={2021}
}

@inproceedings{li2023blip,
  title={Blip-2: Bootstrapping language-image pre-training with frozen image encoders and large language models},
  author={Li, Junnan and Li, Dongxu and Savarese, Silvio and Hoi, Steven},
  booktitle={International conference on machine learning},
  pages={19730--19742},
  year={2023},
  organization={PMLR}
}

@inproceedings{zhou2023esc,
  title={Esc: Exploration with soft commonsense constraints for zero-shot object navigation},
  author={Zhou, Kaiwen and Zheng, Kaizhi and Pryor, Connor and Shen, Yilin and Jin, Hongxia and Getoor, Lise and Wang, Xin Eric},
  booktitle={International Conference on Machine Learning},
  pages={42829--42842},
  year={2023},
  organization={PMLR}
}

@article{yang2023dawn,
  title={The dawn of lmms: Preliminary explorations with gpt-4v (ision)},
  author={Yang, Zhengyuan and Li, Linjie and Lin, Kevin and Wang, Jianfeng and Lin, Chung-Ching and Liu, Zicheng and Wang, Lijuan},
  journal={arXiv preprint arXiv:2309.17421},
  year={2023}
}

@article{chen2019behavioral,
  title={A behavioral approach to visual navigation with graph localization networks},
  author={Chen, Kevin and De Vicente, Juan Pablo and Sepulveda, Gabriel and Xia, Fei and Soto, Alvaro and V{\'a}zquez, Marynel and Savarese, Silvio},
  journal={arXiv preprint arXiv:1903.00445},
  year={2019}
}

@inproceedings{kwon2023renderable,
  title={Renderable neural radiance map for visual navigation},
  author={Kwon, Obin and Park, Jeongho and Oh, Songhwai},
  booktitle={Proceedings of the IEEE/CVF Conference on Computer Vision and Pattern Recognition},
  pages={9099--9108},
  year={2023}
}

@inproceedings{zhu2017target,
  title={Target-driven visual navigation in indoor scenes using deep reinforcement learning},
  author={Zhu, Yuke and Mottaghi, Roozbeh and Kolve, Eric and Lim, Joseph J and Gupta, Abhinav and Fei-Fei, Li and Farhadi, Ali},
  booktitle={2017 IEEE international conference on robotics and automation (ICRA)},
  pages={3357--3364},
  year={2017},
  organization={IEEE}
}

@article{huang2024gamap,
  title={Gamap: Zero-shot object goal navigation with multi-scale geometric-affordance guidance},
  author={Huang, Hao and Hao, Yu and Wen, Congcong and Tzes, Anthony and Fang, Yi and others},
  journal={Advances in Neural Information Processing Systems},
  volume={37},
  pages={39386--39408},
  year={2024}
}

@article{chen2023not,
  title={How to not train your dragon: Training-free embodied object goal navigation with semantic frontiers},
  author={Chen, Junting and Li, Guohao and Kumar, Suryansh and Ghanem, Bernard and Yu, Fisher},
  journal={arXiv preprint arXiv:2305.16925},
  year={2023}
}

@inproceedings{zhang2024imagine,
  title={Imagine before go: Self-supervised generative map for object goal navigation},
  author={Zhang, Sixian and Yu, Xinyao and Song, Xinhang and Wang, Xiaohan and Jiang, Shuqiang},
  booktitle={Proceedings of the IEEE/CVF Conference on Computer Vision and Pattern Recognition},
  pages={16414--16425},
  year={2024}
}

@inproceedings{cao2025cognav,
  title={Cognav: Cognitive process modeling for object goal navigation with llms},
  author={Cao, Yihan and Zhang, Jiazhao and Yu, Zhinan and Liu, Shuzhen and Qin, Zheng and Zou, Qin and Du, Bo and Xu, Kai},
  booktitle={Proceedings of the IEEE/CVF International Conference on Computer Vision},
  pages={9550--9560},
  year={2025}
}

@inproceedings{katyal2019uncertainty,
  title={Uncertainty-aware occupancy map prediction using generative networks for robot navigation},
  author={Katyal, Kapil and Popek, Katie and Paxton, Chris and Burlina, Phil and Hager, Gregory D},
  booktitle={2019 International Conference on Robotics and Automation (ICRA)},
  pages={5453--5459},
  year={2019},
  organization={IEEE}
}

@article{zhang2025apexnav,
  title={ApexNav: An Adaptive Exploration Strategy for Zero-Shot Object Navigation with Target-centric Semantic Fusion},
  author={Zhang, Mingjie and Du, Yuheng and Wu, Chengkai and Zhou, Jinni and Qi, Zhenchao and Ma, Jun and Zhou, Boyu},
  journal={arXiv preprint arXiv:2504.14478},
  year={2025}
}

@inproceedings{ramakrishnan2021hm3d,
  title={Habitat-Matterport 3D Dataset ({HM}3D): 1000 Large-scale 3D Environments for Embodied {AI}},
  author={Santhosh Kumar Ramakrishnan and Aaron Gokaslan and Erik Wijmans and Oleksandr Maksymets and Alexander Clegg and John M Turner and Eric Undersander and Wojciech Galuba and Andrew Westbury and Angel X Chang and Manolis Savva and Yili Zhao and Dhruv Batra},
  booktitle={Thirty-fifth Conference on Neural Information Processing Systems Datasets and Benchmarks Track (Round 2)},
  year={2021},
  url={https://openreview.net/forum?id=-v4OuqNs5P}
}

@inproceedings{yu2023l3mvn,
  title={L3mvn: Leveraging large language models for visual target navigation},
  author={Yu, Bangguo and Kasaei, Hamidreza and Cao, Ming},
  booktitle={2023 IEEE/RSJ International Conference on Intelligent Robots and Systems (IROS)},
  pages={3554--3560},
  year={2023},
  organization={IEEE}
}

@inproceedings{zhang2024trihelper,
  title={Trihelper: Zero-shot object navigation with dynamic assistance},
  author={Zhang, Lingfeng and Zhang, Qiang and Wang, Hao and Xiao, Erjia and Jiang, Zixuan and Chen, Honglei and Xu, Renjing},
  booktitle={2024 IEEE/RSJ International Conference on Intelligent Robots and Systems (IROS)},
  pages={10035--10042},
  year={2024},
  organization={IEEE}
}

@article{wu2024voronav,
  title={Voronav: Voronoi-based zero-shot object navigation with large language model},
  author={Wu, Pengying and Mu, Yao and Wu, Bingxian and Hou, Yi and Ma, Ji and Zhang, Shanghang and Liu, Chang},
  journal={arXiv preprint arXiv:2401.02695},
  year={2024}
}

@article{zhao2023semantic,
  title={Semantic policy network for zero-shot object goal visual navigation},
  author={Zhao, Qianfan and Zhang, Lu and He, Bin and Liu, Zhiyong},
  journal={IEEE Robotics and Automation Letters},
  volume={8},
  number={11},
  pages={7655--7662},
  year={2023},
  publisher={IEEE}
}

@misc{qwen36_35b_a3b,
  title = {{Qwen3.6-35B-A3B}: Agentic Coding Power, Now Open to All},
  author = {{Qwen Team}},
  month = {April},
  year = {2026},
  url = {https://qwen.ai/blog?id=qwen3.6-35b-a3b}
}

@misc{puig2023habitat3,
      title  = {Habitat 3.0: A Co-Habitat for Humans, Avatars and Robots},
      author = {Xavi Puig and Eric Undersander and Andrew Szot and Mikael Dallaire Cote and Ruslan Partsey and Jimmy Yang and Ruta Desai and Alexander William Clegg and Michal Hlavac and Tiffany Min and Theo Gervet and Vladim\'ir Vondru\v{s} and Vincent-Pierre Berges and John Turner and Oleksandr Maksymets and Zsolt Kira and Mrinal Kalakrishnan and Jitendra Malik and Devendra Singh Chaplot and Unnat Jain and Dhruv Batra and Akshara Rai and Roozbeh Mottaghi},
      year={2023},
      archivePrefix={arXiv},
}

@article{wang2025dreamnav,
  title={Dreamnav: A trajectory-based imaginative framework for zero-shot vision-and-language navigation},
  author={Wang, Yunheng and Fang, Yuetong and Wang, Taowen and Feng, Yixiao and Tan, Yawen and Zhang, Shuning and Liu, Peiran and Ji, Yiding and Xu, Renjing},
  journal={arXiv preprint arXiv:2509.11197},
  year={2025}
}

@article{huang2025vistav2,
  title={VISTAv2: World Imagination for Indoor Vision-and-Language Navigation},
  author={Huang, Yanjia and Jiang, Xianshun and Gao, Xiangbo and Wu, Mingyang and Tu, Zhengzhong},
  journal={arXiv preprint arXiv:2512.00041},
  year={2025}
}

@article{edelsbrunner2003shape,
  title={On the shape of a set of points in the plane},
  author={Edelsbrunner, Herbert and Kirkpatrick, David and Seidel, Raimund},
  journal={IEEE Transactions on information theory},
  volume={29},
  number={4},
  pages={551--559},
  year={2003},
  publisher={IEEE}
}

@article{barber1996quickhull,
  title={The quickhull algorithm for convex hulls},
  author={Barber, C Bradford and Dobkin, David P and Huhdanpaa, Hannu},
  journal={ACM Transactions on Mathematical Software (TOMS)},
  volume={22},
  number={4},
  pages={469--483},
  year={1996},
  publisher={Acm New York, NY, USA}
}

@article{zhang2021safe,
  title={Safe occlusion-aware autonomous driving via game-theoretic active perception},
  author={Zhang, Zixu and Fisac, Jaime F},
  journal={arXiv preprint arXiv:2105.08169},
  year={2021}
}

@article{firoozi2024oa,
  title={Oa-mpc: Occlusion-aware mpc for guaranteed safe robot navigation with unseen dynamic obstacles},
  author={Firoozi, Roya and Mir, Alexandre and Camps, Gadiel Sznaier and Schwager, Mac},
  journal={IEEE Transactions on Control Systems Technology},
  volume={33},
  number={3},
  pages={940--951},
  year={2024},
  publisher={IEEE}
}

@article{axelrod2018provably,
  title={Provably safe robot navigation with obstacle uncertainty},
  author={Axelrod, Brian and Kaelbling, Leslie Pack and Lozano-P{\'e}rez, Tom{\'a}s},
  journal={The International Journal of Robotics Research},
  volume={37},
  number={13-14},
  pages={1760--1774},
  year={2018},
  publisher={SAGE Publications Sage UK: London, England}
}

@article{zhou2026beliefmapnav,
  title={Beliefmapnav: 3d voxel-based belief map for zero-shot object navigation},
  author={Zhou, Zibo and Hu, Yue and Zhang, Lingkai and Li, Zonglin and Chen, Siheng},
  journal={Advances in Neural Information Processing Systems},
  volume={38},
  pages={83008--83036},
  year={2026}
}

@inproceedings{zhao2025imaginenav,
  title={Imaginenav: Prompting vision-language models as embodied navigator through scene imagination},
  author={Zhao, Xinxin and Cai, Wenzhe and Tang, Likun and Wang, Teng},
  booktitle={International Conference on Learning Representations},
  volume={2025},
  pages={94387--94401},
  year={2025}
}

@article{wang2026imaginenav++,
  title={ImagineNav++: Prompting Vision-Language Models as Embodied Navigator through Scene Imagination},
  author={Wang, Teng and Zhao, Xinxin and Cai, Wenzhe and Sun, Changyin},
  journal={IEEE Transactions on Pattern Analysis and Machine Intelligence},
  year={2026},
  publisher={IEEE}
}

\clearpage




\addtocontents{toc}{\protect\setcounter{tocdepth}{1}}

\setcounter{tocdepth}{-1} 
\tableofcontents
\setcounter{tocdepth}{1}  

\renewcommand{\thesection}{\Alph{section}}
\renewcommand{\thesubsection}{\thesection.\arabic{subsection}}
\renewcommand{\thesubsubsection}{\thesubsection.\arabic{subsubsection}}
\newcommand{\ours}{Schr\"odinger's Navigator}

\setcounter{section}{0}
\setcounter{subsection}{0}
\setcounter{subsubsection}{0}

\section*{Supplementary Video}
\addcontentsline{toc}{section}{Supplementary Video}
The supplementary video shows occlusion-heavy HM3D episodes and Go2 demonstrations in the Office, Classroom, and Common Room scenes, including executed trajectories, egocentric observations, target predictions, and trajectory-conditioned imagined views.

\section{Hyperparameters, Runtime, and System Configuration}
\label{app:hyperparams}
Unless otherwise stated, the same parameters are used in HM3D simulation and real-world Go2 deployment. All value maps are normalized to $[0,1]$, and distances are measured in meters.

\subsection{Trajectory-imagination hyperparameters}
The sampler uses three virtual camera trajectories, $\mathcal{V}=\{L,R,U\}$, corresponding to left-bypass, right-bypass, and over-the-top futures. All three trajectories are used for future imagination rather than robot execution.

\begin{table}[h]
\centering
\small
\caption{Geometric parameters for adaptive occluder-aware virtual-camera trajectory sampling.}
\label{tab:traj_params}
\begin{tabular}{lclp{0.46\linewidth}}
\toprule
Parameter & Symbol & Value & Description \\
\midrule
Trajectory ensemble size & $N$ & 3 & Left, right, and over-the-top futures. \\
Camera poses per trajectory & $N_{\mathrm{pose}}$ & 16 & Initial camera poses before interpolation. \\
Nominal radial offset & $\delta$ & 0.20 m & Offset used to define the nominal orbit radius. \\
Minimum orbit radius & $d_c^{\min}$ & 1.0 m & Lower bound on the nominal orbit radius. \\
Nominal orbit radius & $d_c$ & -- & $d_c=\max(r_t+r_{\mathrm{robot}}+\delta,d_c^{\min})$. \\
Angular expansion & $\Delta\theta$ & $20^{\circ}$ & Extends lateral views beyond the observed occluder boundary. \\
Nominal minimum arc extent & $d_v^{\min}$ & 1.0 m & Lower setting for the nominal arc extent; realized trajectories may be shorter. \\
Maximum trajectory length & $d_v^{\max}$ & 3.0 m & Upper bound on the sampled camera path length. \\
Lateral edge-gaze weight & $\gamma_{\mathrm{edge}}$ & 0.35 & Interpolates the lateral gaze target toward the corresponding boundary side. \\
\bottomrule
\end{tabular}
\end{table}

\subsection{FAVM, perception, and runtime settings}
\begin{table}[h]
\centering
\small
\caption{Default coefficients for FAVM construction and fusion.}
\label{tab:favm_params}
\begin{tabular}{lclp{0.42\linewidth}}
\toprule
Parameter & Symbol & Value & Description \\
\midrule
Observation/future fusion balance & $\beta$ & 0.60 & Balances $\bar m_t$ and $m_t^{\mathrm{FA}}$. \\
Semantic score weight & $\alpha_{\mathrm{sem}}$ & 1.0 & Weight for target relevance $S_t(p)$. \\
Exploration score weight & $\alpha_{\mathrm{exp}}$ & 1.0 & Weight for expected reveal $E_t(p)$. \\
Risk score weight & $\alpha_{\mathrm{risk}}$ & 1.0 & Weight for traversal risk $R_t(p)$. \\
Imagined semantic discount & $\lambda_{\mathrm{img}}$ & 0.50 & Discount for imagined target cues. \\
Semantic-transfer depth tolerance & $\tau_{d,i}^{(m)}$ & Adaptive & Radius- and depth-dependent; see Appendix~C. \\
\bottomrule
\end{tabular}
\end{table}

We use Qwen3.6-35B-A3B \citep{qwen36_35b_a3b} for instruction parsing and direction judgment, FlashWorld \citep{li2025flashworld} for trajectory-conditioned 3D imagination, and GLEE \citep{wu2024general} for open-vocabulary semantic segmentation. FlashWorld uses $480\times704$ input resolution, 24 conditioning camera poses interpolated from the 16 initial poses, and sparse triggering. Semantic transfer uses three rendered views; optional videos are exported at 15 fps.

\subsection{Runtime characteristics}
\textbf{Trig./Ep.} denotes FlashWorld invocations per episode; \textbf{Lat./Trig.} denotes per-trigger world-model service time; \textbf{Wait Ratio} is the fraction of episode time waiting for fresh future evidence; and \textbf{Waypoint Change} measures the fraction of navigation decisions in which the selected waypoint differs from that selected using the observation-only map.
In HM3D simulation, FlashWorld is triggered at a fixed interval of 50 observation updates. In real-world execution it runs asynchronously while the local planner continues executing the current waypoint. In our experiments, $N=1/3/5$ futures take about $3.0/9.0/15.0$s per trigger. The full model triggers 4.24 times per episode, with a 20.1\% waiting ratio and a 46.8\% waypoint-change rate.

\section{Occluder-Aware Trajectory Sampling}
\label{app:trajectory_sampling}
The sampler generates a compact set of virtual camera trajectories around the dominant forward occluder rather than densely enumerating actions. It is designed to cover complementary lateral and vertical reveal directions.

\subsection{Point cloud preprocessing and occluder selection}
Given RGB-D observation $(I_t,D_t)$, intrinsics $K$, and pose $q_t$, valid depth pixels are back-projected as
\begin{equation}
    x(p)=D_t(p)K^{-1}\bar p,\qquad \bar p=[u,v,1]^\top .
\end{equation}
Ground points are removed, and the remaining non-ground points are clustered into $\mathcal{C}_t=\{C_1,\ldots,C_M\}$. A cluster is a forward candidate if
\begin{equation}
    C_j \in \mathcal{C}^{\mathrm{fwd}}_t
    \Longleftrightarrow
    \left|\operatorname{wrap}\left(\angle(\bar x_j-x_t)-\psi_t\right)\right|\leq\theta_{\mathrm{fwd}}.
\end{equation}
We rank each forward cluster by normalized projected area, inverse distance, and heading alignment:
\begin{equation}
    \rho_j=\overline{A}_j+(1-\overline d_j)+\overline h_j,
    \qquad
    h_j=\max\{0,\cos(\angle(\bar x_j-x_t)-\psi_t)\},
\end{equation}
and select $O_t=\arg\max_{C_j\in\mathcal{C}^{\mathrm{fwd}}_t}\rho_j$.

\subsection{Boundary, orbit center, and angular span}
The selected occluder $O_t$ is projected onto the ground plane, and its boundary $B_t$ is extracted using a convex hull \citep{barber1996quickhull} or an alpha-shape contour \citep{edelsbrunner2003shape}. We define
\begin{equation}
    c_t=\frac{1}{|B_t|}\sum_{b\in B_t}b,
    \qquad
    r_t=\max_{b\in B_t}\|b-c_t\|_2,
    \qquad
    d_c=\max(r_t+r_{\mathrm{robot}}+\delta,d_c^{\min}).
\end{equation}
For each boundary point $b$, let $\theta(b)=\operatorname{atan2}(b_y-c_{t,y},b_x-c_{t,x})$. After unwrapping, the occluder span is $[\theta_{\min},\theta_{\max}]$. Lateral target angles and arc lengths are
\begin{equation}
    \theta_L=\theta_{\max}+\Delta\theta,
    \quad
    \theta_R=\theta_{\min}-\Delta\theta,
    \quad
    d_v^{(v)}=\operatorname{clamp}\!\big(d_c|\operatorname{wrap}(\theta_v-\theta_0)|,d_v^{\min},d_v^{\max}\big),
\end{equation}
where $v\in\{L,R\}$ and $\theta_0$ is the current camera azimuth around $c_t$. The clamped length specifies a nominal arc extent; the realized start-anchored camera trajectory may be shorter than $d_v^{\min}$.

\subsection{Trajectory instantiation and filtering}
Based on the occluder geometry, the sampler generates continuous left,
right, and overhead virtual camera trajectories from the current camera
pose. Each trajectory is represented as
\begin{equation}
    \tau_t^{(v)}
    =\{(p_k^{(v)},R_k^{(v)})\}_{k=0}^{N_{\mathrm{pose}}-1},
    \quad v\in\{L,R,U\}.
\end{equation}
The lateral trajectories follow occluder-aware arcs with angular extents
determined by the observed occluder boundary.

For each lateral trajectory, the horizontal camera gaze target is defined as
\begin{equation}
    g_{t,xy}^{(v)}
    =(1-\gamma_{\mathrm{edge}})c_{t,xy}
    +\gamma_{\mathrm{edge}}e_{t,xy}^{(v)},
    \quad v\in\{L,R\},
\end{equation}
where $e_{t,xy}^{(v)}$ denotes the corresponding observed boundary-side
gaze target and $\gamma_{\mathrm{edge}}=0.35$. Camera orientations are smoothly interpolated toward the
target viewing directions using spherical linear interpolation (SLERP).

For the overhead trajectory, the camera faces the central direction
of the observed occluder boundary, while its height varies smoothly as
\begin{equation}
    z_k^{(U)}
    =z_{\mathrm{cam}}+
    h_{\max}\sin\left(\frac{\pi k}{N_{\mathrm{pose}}-1}\right).
\end{equation}

The sampler validates the generated camera trajectories for pose validity, continuity, and compatibility with the world-model input. If no reliable forward occluder is identified or the adaptive trajectories fail to meet the generation requirements, the sampler falls back to the predefined fixed left, right, and overhead virtual camera trajectories.

\section{3DGS Alignment, Semantic Label Transfer, and Scene Fusion}
\label{app:scene_fusion}
This section summarizes how trajectory-conditioned 3DGS futures are converted into a planning representation.

\subsection{Representation and coordinate alignment}
Each Gaussian is stored compactly as
\begin{equation}
    g_i=[x_i,y_i,z_i,c_i^r,c_i^g,c_i^b,\mathrm{rad}_i,\mathrm{opa}_i,\ell_i],
\end{equation}
where $(x_i,y_i,z_i)$ is the center, $(c_i^r,c_i^g,c_i^b)$ is color, $\mathrm{rad}_i$ and $\mathrm{opa}_i$ are simplified size and opacity, and $\ell_i$ is the semantic label.
Before generation, the current pose and sampled trajectory are transformed into a local camera-centered frame $\hat W$:
\begin{equation}
    \hat q_t=T_{\hat W\leftarrow W}(q_t),\qquad
    \hat\tau_t^{(v)}=T_{\hat W\leftarrow W}\tau_t^{(v)},
    \qquad
    \hat G_t^{(v)}=\Phi_\theta(I_t,\hat q_t,\hat\tau_t^{(v)}).
\end{equation}
After scale alignment, Gaussian centers are transformed back to the world frame:
\begin{equation}
    \tilde\mu_i^{(v)}=R_t(s_t^{(v)}\hat\mu_i^{(v)})+t_t.
\end{equation}
The compact imagined primitives use a fixed radius of $0.03\,\mathrm{m}$ and opacity of $1$ after filtering.

\subsection{Depth scale alignment and semantic transfer}
For trajectory $v$, we render $\hat D_t^{(v)}$ from $\hat G_t^{(v)}$ at the current viewpoint and estimate scale by the median depth ratio:
\begin{equation}
    \Omega_t^{(v)}=\{p\mid 0.1<D_t(p)<5.0,\hat D_t^{(v)}(p)>10^{-4}\},
    \qquad
    s_t^{(v)}=\operatorname{median}_{p\in\Omega_t^{(v)}}\frac{D_t(p)}{\hat D_t^{(v)}(p)}.
\end{equation}
The observed depth is center-cropped and resized using nearest-neighbor interpolation before alignment. At least 100 valid pixels are required; otherwise, alignment raises an error.
For semantic transfer, each available view $m$ has a 2D semantic map $Y_t^{(m)}(p)$. A projected primitive receives a valid vote only if it is inside the image and depth-consistent:
\begin{equation}
    \begin{aligned}
    p_i^{(m)}&=\pi(KT_{C_m\leftarrow W}\mu_i),\\
    \left|z_i^{(m)}-D_{\mathrm{rend}}^{(m)}(p_i^{(m)})\right|
        &\leq\tau_{d,i}^{(m)},\\
    \tau_{d,i}^{(m)}
        &=\max\left\{2\,\mathrm{rad}_i,\,
        \frac{D_{\mathrm{rend}}^{(m)}(p_i^{(m)})}
        {f_{\mathrm{pix}}^{(m)}}\right\}.
    \end{aligned}
\end{equation}
Here $z_i^{(m)}$ is the positive camera-frame depth of the Gaussian center, and $f_{\mathrm{pix}}^{(m)}$ is the smaller of the horizontal and vertical focal lengths in pixels. Only finite, positive rendered depths are used.
Its final label is obtained by majority voting over valid semantic labels:
\begin{equation}
    \ell_i=\arg\max_c\sum_{m\in\mathcal{M}_i}\mathbf{1}[Y_t^{(m)}(p_i^{(m)})=c].
\end{equation}
Segmentation confidence is thresholded at $0.3$ before voting; primitives with no valid votes remain unlabeled.

\subsection{Scene fusion}
The aligned imagined primitives are first collected as
\begin{equation}
    G_t^{\mathrm{img}}=\bigcup_{v\in\mathcal{V}}\tilde G_t^{(v)},
    \qquad
    G_t^{\mathrm{fuse}}=\operatorname{Merge}(G_t^{\mathrm{obs}}\cup G_t^{\mathrm{img}}).
\end{equation}
Here $\operatorname{Merge}$ denotes the joint use of separately stored observed and imagined representations, rather than confidence-weighted parameter averaging. The source indicator $\sigma_g\in\{\mathrm{obs},\mathrm{img}\}$ is determined by buffer membership. Imagined trajectory outputs are concatenated after per-trajectory filtering and voxel downsampling, while the observed map is accumulated and downsampled separately. Voxel downsampling retains one representative primitive per occupied voxel.

\begin{figure}[t]
    \centering
    \includegraphics[width=0.96\linewidth]{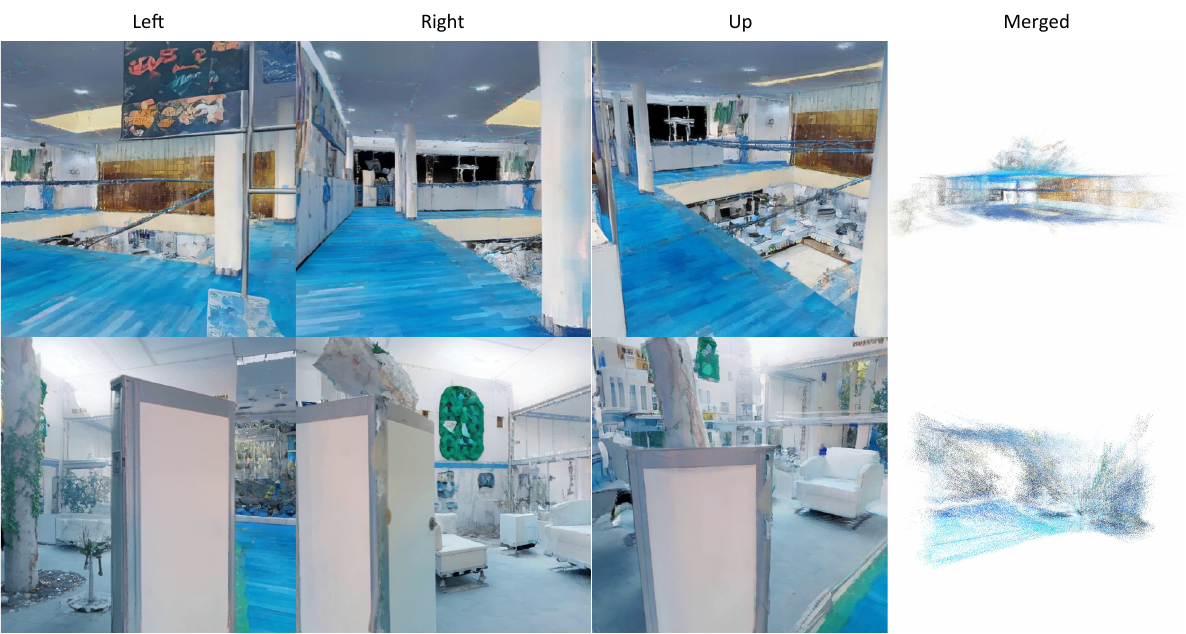}
    \caption{Trajectory-conditioned 3DGS generation and fusion. Each row shows left-bypass, right-bypass, over-the-top, and merged futures.}
    \label{fig:app_3dgs_fusion}
\end{figure}

\section{Future-Aware Value Map and Observation-Based Component Maps}
\label{app:favm_maps}
FAVM is evaluated on candidate waypoints using $G_t^{\mathrm{fuse}}$, while $\sigma_g$ separates observed evidence from imagined hypotheses.

\subsection{FAVM components}
Let $\mathcal{P}_t=\{\mu_g\mid g\in\mathcal{G}_{\rm nav}\}$ denote candidate navigation positions given by the centers of navigable Gaussians.
All scalar maps are normalized over $\mathcal{P}_t$ by
\begin{equation}
    \mathcal{N}(x)(p)=\frac{x(p)-\min_{q\in\mathcal{P}_t}x(q)}{\max_{q\in\mathcal{P}_t}x(q)-\min_{q\in\mathcal{P}_t}x(q)+\epsilon}.
\end{equation}
For target category $y^\star$, observed and imagined target sets are
\begin{equation}
    T_t^{\mathrm{real}}=\{g\in G_t^{\mathrm{fuse}}\mid \ell_g=y^\star,\sigma_g=\mathrm{obs}\},\quad
    T_t^{\mathrm{hyp}}=\{g\in G_t^{\mathrm{fuse}}\mid \ell_g=y^\star,\sigma_g=\mathrm{img}\}.
\end{equation}
The semantic score discounts imagined target cues:
\begin{equation}
    \tilde S_t(p)=e^{-d_t^{\mathrm{real}}(p)/\tau_{\mathrm{sem}}}
    +\lambda_{\mathrm{img}}e^{-d_t^{\mathrm{hyp}}(p)/\tau_{\mathrm{sem}}},
    \quad S_t=\mathcal{N}(\tilde S_t).
\end{equation}
Here $d_t^{a}(p)=\min_{g\in T_t^{a}}\|p-\mu_g\|_2$ for
$a\in\{\mathrm{real},\mathrm{hyp}\}$,
$\tau_{\mathrm{sem}}=1.0\,\mathrm{m}$, and
$\lambda_{\mathrm{img}}=0.5$.
An empty target set contributes zero.
At each imagination update, $F_t^{\mathrm{new}}$ is constructed from imagined primitives lying less than $0.20\,\mathrm{m}$ from the estimated floor height and more than $0.15\,\mathrm{m}$ from the observed navigable map, followed by voxel downsampling. The expected reveal score is approximated by local coverage:
\begin{equation}
    \tilde E_t(p)=\sum_{g\in F_t^{\mathrm{new}}}
    \mathbf{1}[\|p-\mu_g\|_2\le r_{\mathrm{vis}}],
    \quad E_t=\mathcal{N}(\tilde E_t).
\end{equation}
Here $r_{\mathrm{vis}}=0.60\,\mathrm{m}$.
For risk estimation, we extract a set of imagined occupied voxels
$H_t^{\mathrm{img}}$ through depth-consistency checks, geometric filtering,
and spatial deduplication. Let $d_{\mathrm{XY}}(h,p)$ denote the shortest
ground-plane distance from voxel $h$ to the line segment connecting the
current camera position and candidate waypoint $p$. The geometric risk
score is computed as
\begin{equation}
    \tilde R_t(p)
    =\sum_{h\in H_t^{\mathrm{img}}}
    \exp\left(-\frac{d_{\mathrm{XY}}(h,p)^2}{2r_{\mathrm{path}}^2}\right),
    \qquad R_t=\mathcal N(\tilde R_t).
\end{equation}
We set $r_{\mathrm{path}}=0.35\,\mathrm{m}$. When no valid risk evidence is
available, $R_t$ is set to zero.
The Future-Aware Value Map is
\begin{equation}
    m_t^{\mathrm{FA}}(p)=\mathcal{N}\left(\alpha_{\mathrm{sem}}S_t(p)+\alpha_{\mathrm{exp}}E_t(p)-\alpha_{\mathrm{risk}}R_t(p)\right),\qquad p\in\mathcal{P}_t.
\end{equation}

\subsection{Observation-based map and navigation target}
The observation-based map is
\begin{equation}
    \bar m_t(p)=\mathcal{N}\left(m_t^{\mathrm{act}}(p)+m_t^{\mathrm{sem}}(p)+m_t^{\mathrm{traj}}(p)+m_t^{\mathrm{heu}}(p)\right).
\end{equation}
Here $m_t^{\mathrm{sem}}$ favors proximity to currently observed landmarks, $m_t^{\mathrm{act}}$ encodes the next high-level action sector or frontier preference, $m_t^{\mathrm{traj}}$ discourages revisits using trajectory history, and $m_t^{\mathrm{heu}}$ projects the multimodal planner's predicted direction onto the map. The final affordance map and navigation target are
\begin{equation}
    m_t^{\mathrm{aff}}(p)=\beta\bar m_t(p)+(1-\beta)m_t^{\mathrm{FA}}(p),
    \qquad
    p_t^\star=\arg\max_{p\in\mathcal{P}_t}m_t^{\mathrm{aff}}(p).
\end{equation}
The selected navigation target is used by the local planner to generate executable waypoints. Between triggers, the robot uses the latest observation-based map and the most recent valid FAVM.

\begin{figure}[t]
    \centering
    \includegraphics[width=0.96\linewidth]{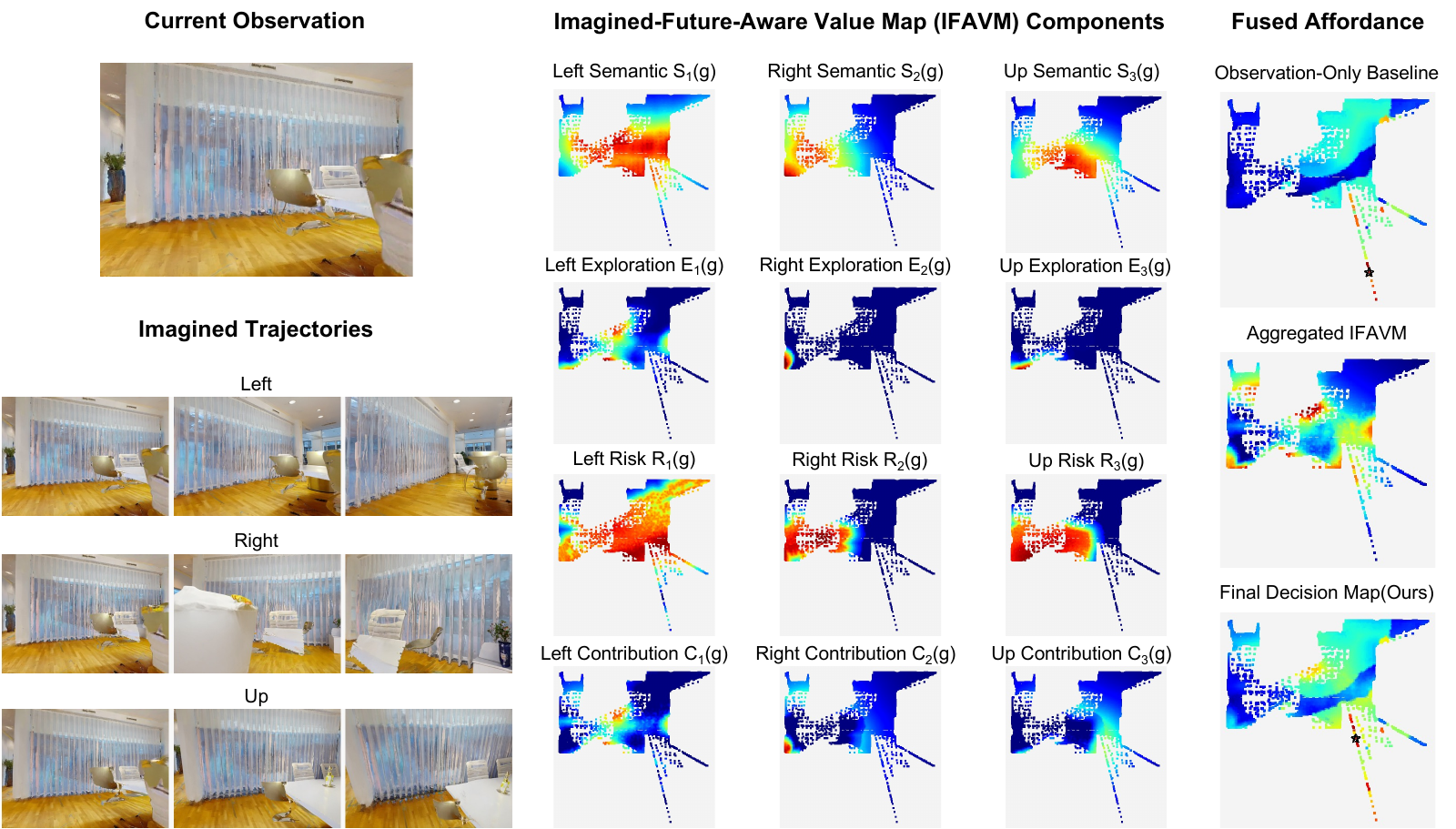}
    \caption{Future-Aware Value Map construction. The figure shows the current observation, left/right/up imagined futures, per-trajectory semantic, exploration, risk, and contribution maps, and the final fused affordance map. Warm colors indicate higher utility, except in risk maps where they indicate higher penalty; the in-figure label IFAVM denotes the FAVM decomposition.}
    \label{fig:app_favm_components}
\end{figure}

For visualization, the per-trajectory contribution is
\begin{equation}
    C_t^{(v)}(p)=\mathcal{N}\left(\alpha_{\mathrm{sem}}S_t^{(v)}(p)+\alpha_{\mathrm{exp}}E_t^{(v)}(p)-\alpha_{\mathrm{risk}}R_t^{(v)}(p)\right).
\end{equation}
Figure~\ref{fig:app_favm_components} shows that FAVM shifts value from visible frontiers toward regions that balance target relevance, expected reveal, and traversal safety.

\FloatBarrier

\section{Baselines, Evaluation Protocol, and Metrics}
\label{app:baselines}

\subsection{Simulation benchmark and metrics}

We evaluate on HM3D within Habitat. Following the ObjectNav protocol, each episode is capped at a maximum of 500 steps and is considered successful if the agent stops within 1.0 m of an instance of the target category. We report Success Rate (SR), Success weighted by Path Length (SPL), and Distance to Goal (DTG). SR measures the fraction of successful episodes. SPL rewards successful episodes with shorter paths. DTG measures the final geodesic distance from the agent to the target object.
In HM3D simulation, the selected waypoints are executed using Habitat's \texttt{ShortestPathFollower} with discrete navigation actions.

We compare against four groups of baselines.

\paragraph{Task-trained approaches.}
ZSON \citep{majumdar2022zson}, PixNav \citep{cai2024bridging}, SPNet \citep{zhao2023semantic}, and SGM \citep{zhang2024imagine} rely on task-specific training or learned navigation policies. These methods can perform well in their training domains but are less aligned with the training-free zero-shot setting targeted by our method.

\paragraph{Zero-shot geometric and semantic explorers.}
ESC \citep{zhou2023esc}, L3MVN \citep{yu2023l3mvn}, TriHelper \citep{zhang2024trihelper}, VoroNav \citep{wu2024voronav}, GAMap \citep{huang2024gamap}, and BeliefMapNav \citep{zhou2026beliefmapnav} use geometric frontiers, semantic maps, language priors, or belief maps to guide exploration without task-specific fine-tuning. They improve semantic exploration but typically estimate value from observed frontiers or a single belief state rather than from an ensemble of trajectory-conditioned future 3D scenes.

\paragraph{LLM/VLM-based planners.}
VLFM \citep{yokoyama2024vlfm} and InstructNav \citep{long2024instructnav} use vision-language or language-model reasoning to construct navigation guidance. For real-world comparison, we use a Qwen3.6-35B-A3B-based implementation of InstructNav-style planning to align the high-level reasoning capability with our system. This baseline uses the same RGB-D observations and local controller where applicable, but it does not use trajectory-conditioned 3D future imagination or FAVM.

\paragraph{Imagination-based navigators.}
ImagineNav \citep{zhao2025imaginenav} and ImagineNav++ \citep{wang2026imaginenav++} use scene imagination to assist embodied navigation. Unlike these methods, Schr\"odinger's Navigator imagines 3DGS futures conditioned on multiple candidate trajectories and fuses the resulting semantic, exploratory, and risk evidence into a spatial FAVM for zero-shot ObjectNav.

\begin{table}[t]
\centering
\small
\caption{HM3D success rates stratified by occlusion severity. $\Delta$SR denotes Ours minus ImagineNav++.}
\label{tab:occlusion_severity}
\begin{tabular}{lccc}
\toprule
Occlusion & ImagineNav++ SR $\uparrow$ & Ours SR $\uparrow$ & $\Delta$SR $\uparrow$ \\
\midrule
Low    & 0.682 & 0.720 & $+0.038$ \\
Medium & 0.629 & 0.674 & $+0.045$ \\
High   & 0.564 & 0.652 & $+0.088$ \\
\bottomrule
\end{tabular}
\end{table}

As occlusion severity increases, both methods exhibit lower success rates, while the advantage of our method becomes more pronounced. Our method outperforms ImagineNav++ by 3.8, 4.5, and 8.8 percentage points under low, medium, and high occlusion, respectively.

\subsection{Real-world protocol}

Real-world experiments are conducted on a Unitree Go2 quadruped equipped with an RGB-D camera in three indoor environments: Office, Classroom, and Common Room. We evaluate two settings. In Static Object Search, the target is a static object in a standard indoor scene. In Unknown-Risk Scenarios, the scene includes severe occlusion or blind spots that may hide targets or risky regions from the initial viewpoint. Each scene contains 10 trials per setting, and success is reported as Success / Total Trials.

\section{Additional Qualitative Results and Failure Analysis}
\label{app:qualitative_failure}
In qualitative overlays, \textbf{Flag} denotes the visual target-verification signal and \textbf{DTS} denotes the logger's distance-to-success value; main quantitative results use DTG.

\begin{figure}[H]
    \centering
    \includegraphics[width=0.98\linewidth]{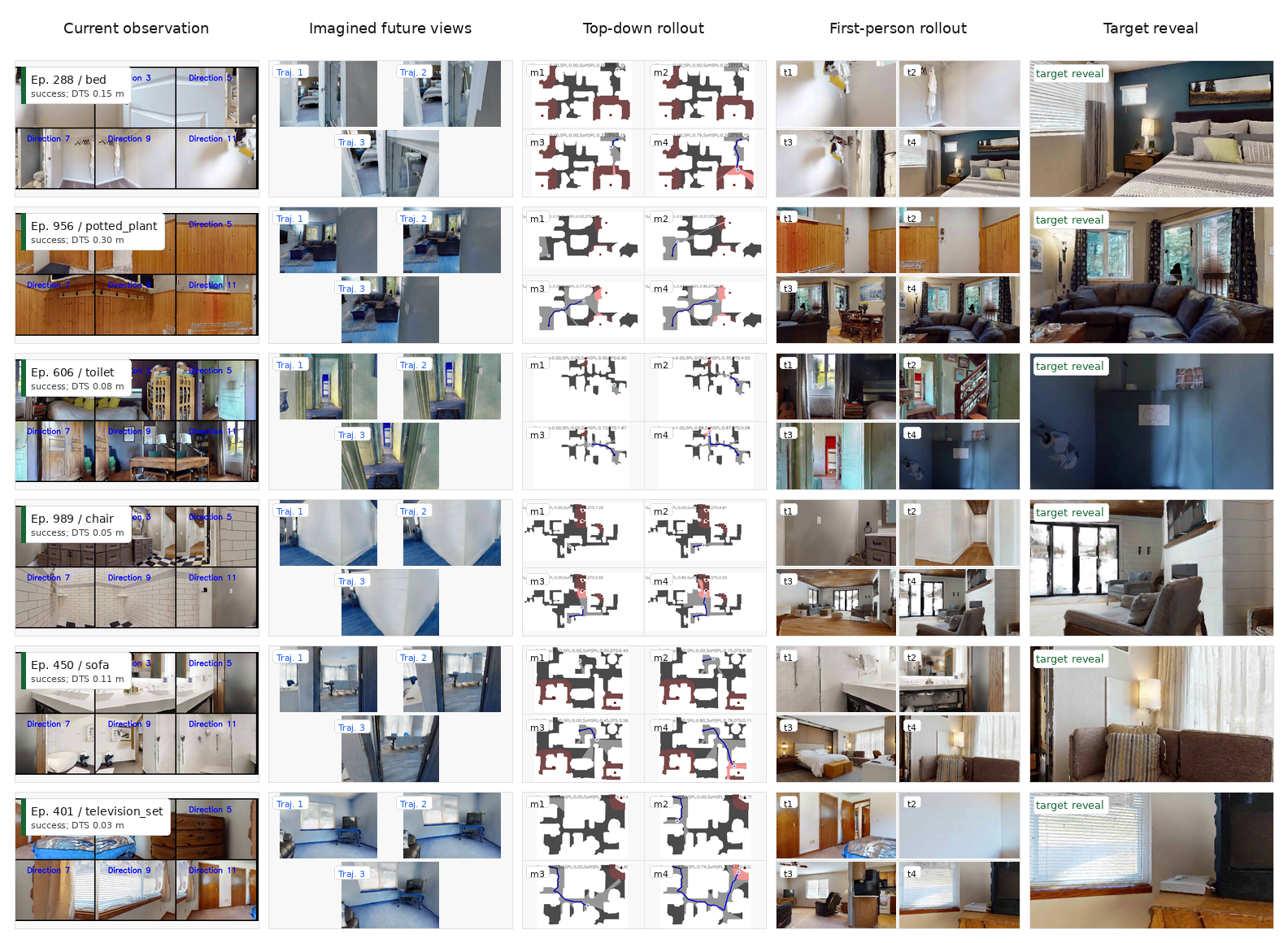}
    \caption{Successful occlusion-heavy simulation rollouts. Each row shows the current observation, imagined future views, top-down checkpoints, first-person rollout frames, and the final target reveal.}
    \label{fig:app_success_rollouts}
\end{figure}

\paragraph{Successful rollouts.}
Figure~\ref{fig:app_success_rollouts} shows cases where imagined futures guide the agent toward viewpoints that reveal hidden targets across bed, potted plant, toilet, chair, sofa, and television set episodes.

\begin{figure}[H]
    \centering
    \includegraphics[width=0.98\linewidth]{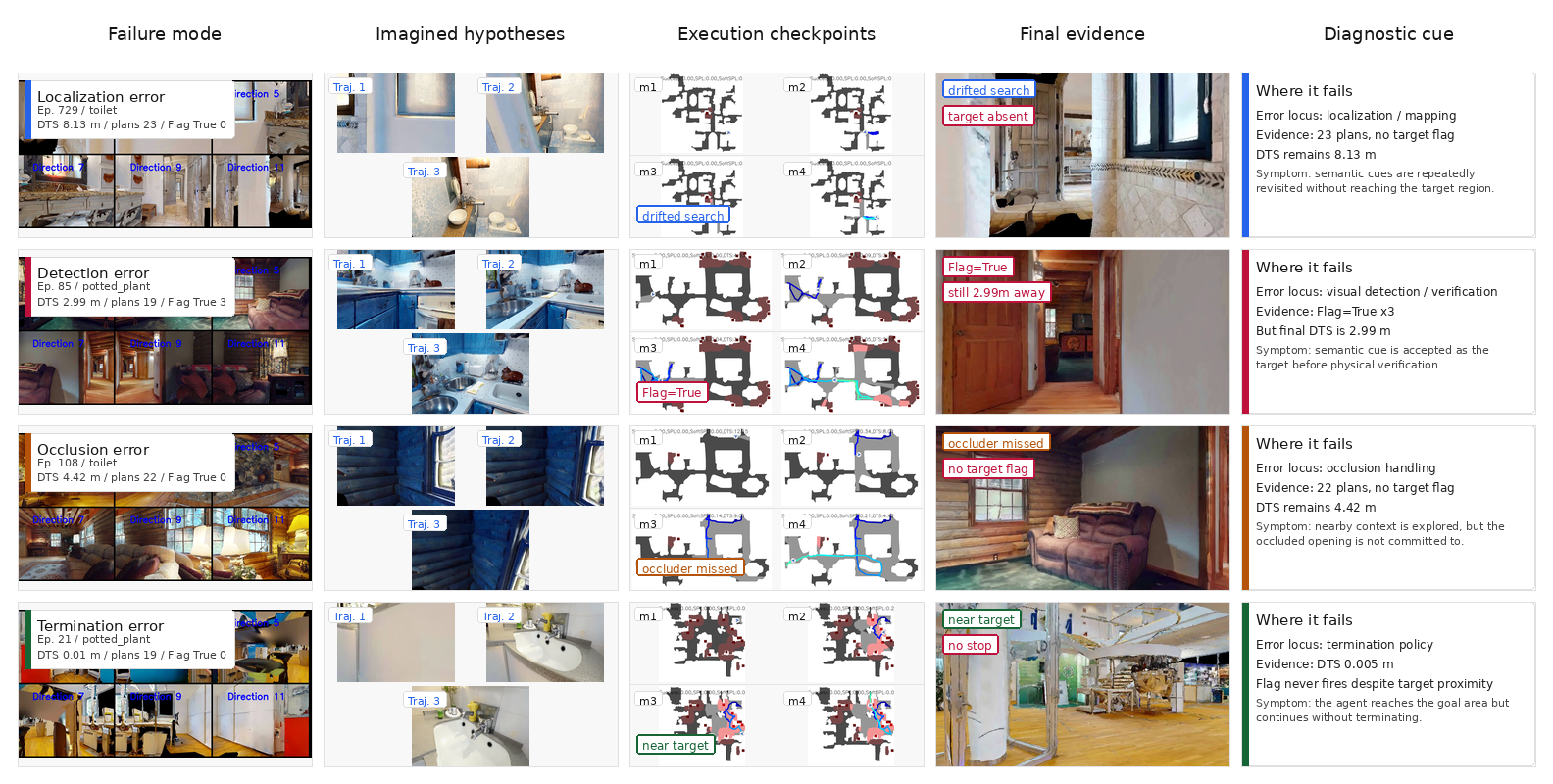}
    \caption{End-to-end failure modes. Rows show localization, detection, occlusion-handling, and termination errors with imagined hypotheses, execution checkpoints, final evidence, and diagnostic cues.}
    \label{fig:app_failure_cases}
\end{figure}

\paragraph{End-to-end failures.}
Figure~\ref{fig:app_failure_cases} summarizes localization, detection, occlusion-handling, and termination failures, which typically arise from compounding perception, mapping, planning, and stopping errors.

\begin{figure}[H]
    \centering
    \includegraphics[width=0.98\linewidth]{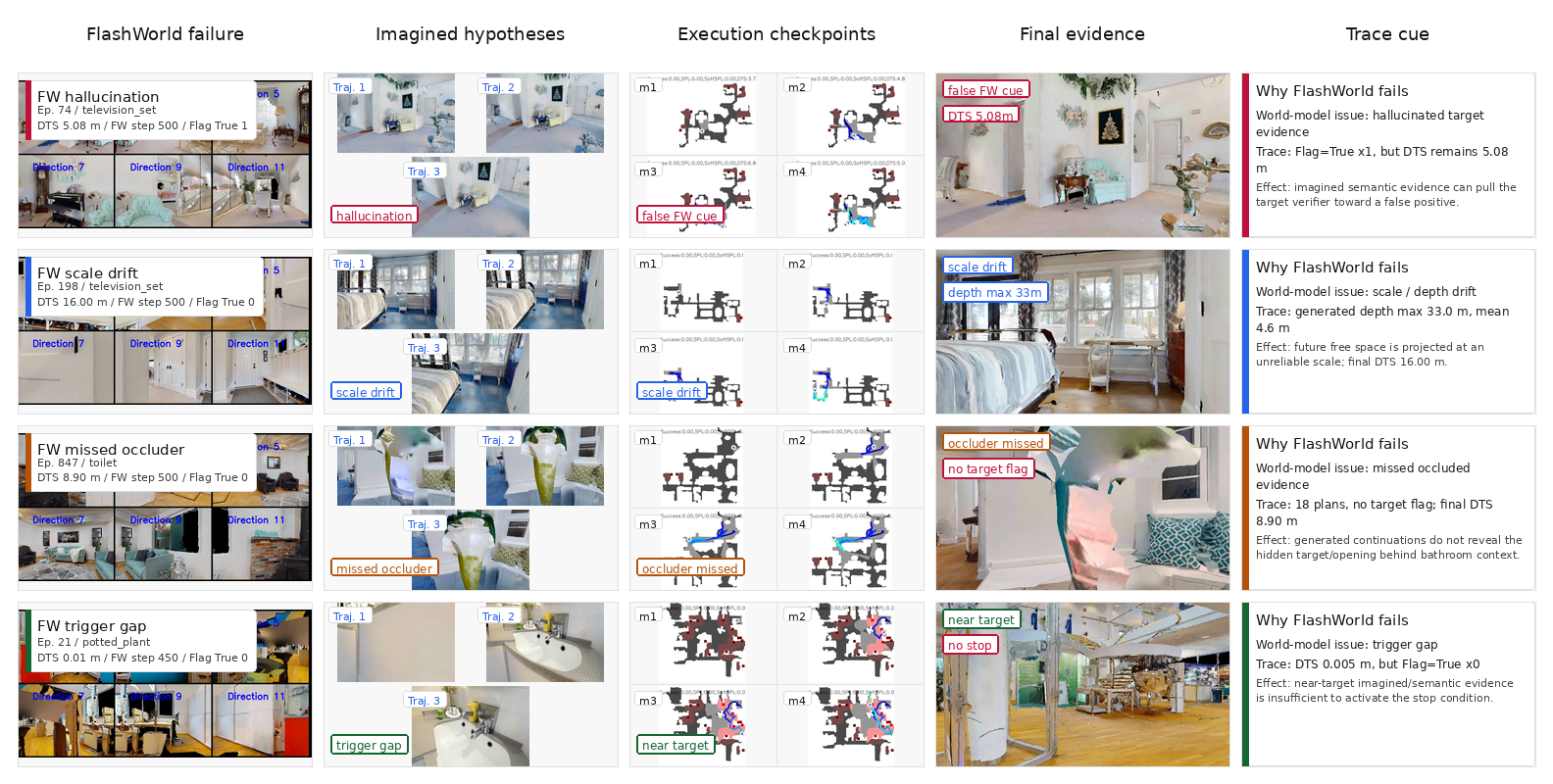}
    \caption{FlashWorld-related failure modes. Rows show hallucinated target evidence, scale/depth drift, missed occluded evidence, and trigger gaps.}
    \label{fig:app_flashworld_failures}
\end{figure}

\paragraph{World-model failures.}
Figure~\ref{fig:app_flashworld_failures} isolates hallucinated target evidence, scale/depth drift, missed occluded evidence, and trigger gaps. These cases motivate discounting imagined semantics, penalizing geometrically supported occupied regions along candidate paths, and fusing future evidence with observation-based maps.

\section{Real-World Deployment Details}
\label{app:deployment}
\subsection{Hardware, services, and safety}
Real-world experiments use a Unitree Go2 quadruped with an Intel RealSense D435i RGB-D camera. The onboard Jetson Orin maintains robot-state feedback and velocity control, while RGB-D observations are forwarded through FastAPI to a remote GPU server for perception, future imagination, and FAVM waypoint selection. The asynchronous split keeps high-frequency control independent of the slower world-model service.

\begin{figure}[H]
    \centering
    \includegraphics[width=0.66\linewidth]{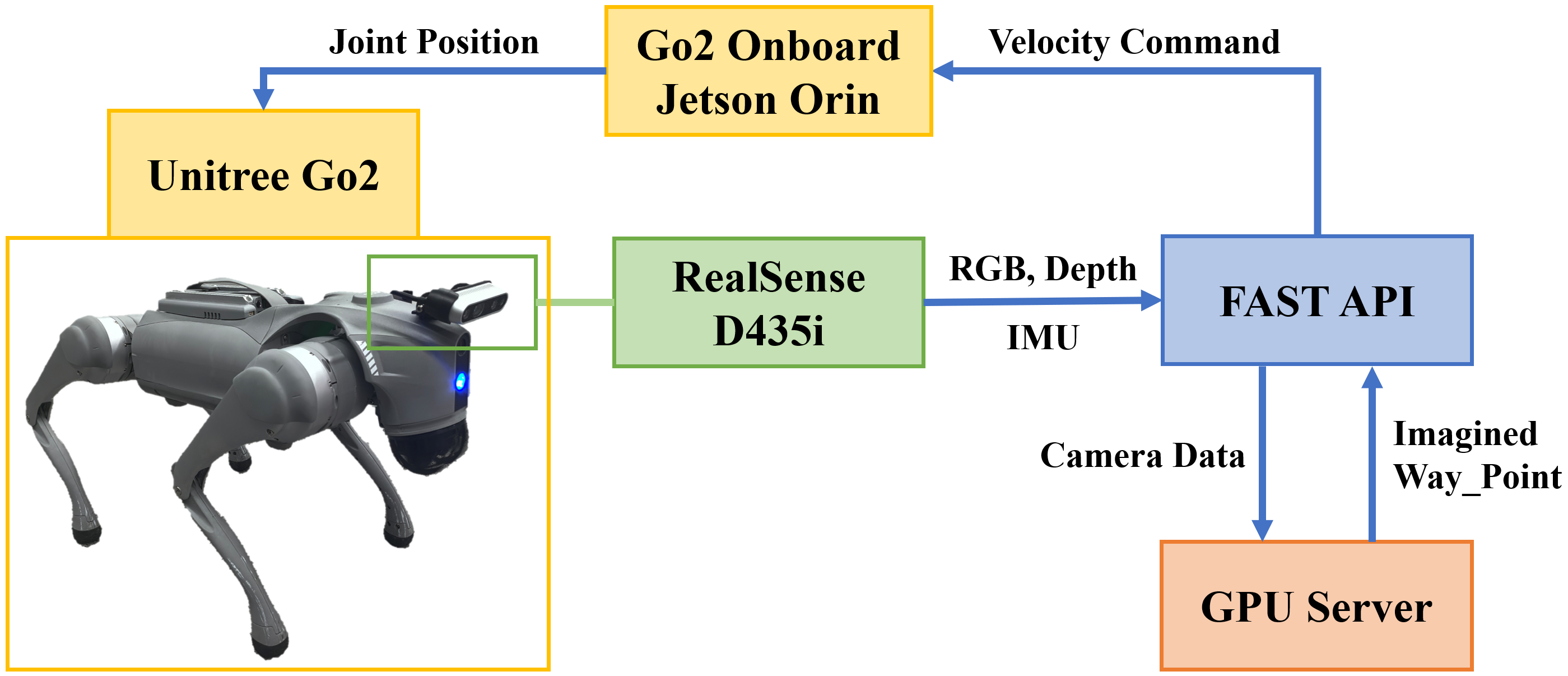}
    \caption{Real-world deployment architecture for Go2 sensing, FastAPI communication, remote GPU inference, and velocity-command execution.}
    \label{fig:app_hardware_architecture}
\end{figure}

\begin{table}[h]
\centering
\small
\caption{Compute and service configuration.}
\label{tab:compute_config}
\begin{tabular}{ll}
\toprule
Item & Configuration \\
\midrule
Robot platform & Unitree Go2 quadruped \\
Onboard computer & Go2 onboard Jetson Orin \\
RGB-D sensor & Intel RealSense D435i \\
Simulation compute & Four NVIDIA RTX 4090 GPUs (parallel evaluation) \\
Real-world inference GPU & One NVIDIA H800 GPU on a remote server \\
World-model service & FlashWorld on the remote GPU server via FastAPI \\
High-level reasoning & Qwen3.6-35B-A3B for instruction and direction judgment \\
Control interface & FastAPI waypoint service with velocity-command output \\
\bottomrule
\end{tabular}
\end{table}

For HM3D simulation, four RTX 4090 GPUs support parallel episode evaluation and shared model inference, increasing evaluation throughput without changing the per-episode navigation protocol.

Each control cycle updates the local point cloud and semantic/navigable maps, parses action-landmark guidance, triggers future imagination when needed, fuses observation-based maps with FAVM, and sends the selected waypoint to the local planner. In Static Object Search, targets are common indoor objects in plausible locations; in Unknown-Risk Scenarios, targets or risky regions are partially hidden by furniture, dividers, desks, or plants. The robot executes only locally collision-checked waypoints. Imagined futures modify affordance scores but never directly command motion, and all real-world trials use conservative speed limits with manual supervision.

\FloatBarrier


\end{document}